\documentclass[english,ruled]{article}
\usepackage[T1]{fontenc}
\usepackage[latin9]{inputenc}
\usepackage{verbatim}
\usepackage[algo2e,ruled]{algorithm2e}
\usepackage{amsmath}
\usepackage{amsthm}
\usepackage{amssymb}
\usepackage{graphicx}
\usepackage{xcolor}
\usepackage{enumitem}
\makeatletter
\usepackage[toc,page,header]{appendix}
\usepackage{minitoc}
\usepackage[accepted]{icml2025}
\usepackage{comment}
\usepackage{ifthen}
\newboolean{doublecolumn}
\setboolean{doublecolumn}{true}
\usepackage{multirow}
\usepackage{bm}

\renewcommand \thepart{}
\renewcommand \partname{}

\usepackage{babel}

\usepackage{hyperref}
\usepackage{url}
\usepackage{booktabs}
\usepackage{amsfonts}
\usepackage{nicefrac}
\usepackage{authblk}
\usepackage{optidef}
\usepackage{array}
\hypersetup{
    colorlinks=true, 
    linkcolor=red, 
    citecolor=blue, 
    urlcolor=magenta 
}

\makeatother

\usepackage{subcaption}
\usepackage{color-edits}
\addauthor{ev}{purple}

\usepackage[capitalize,noabbrev]{cleveref}

\theoremstyle{plain}
\newtheorem{thm}{Theorem}[section]
\newtheorem{prop}[thm]{Proposition}
\newtheorem{lem}[thm]{Lemma}

\theoremstyle{definition}

\newtheorem{ass}[thm]{Assumption}

\newtheoremstyle{remark}
  {\topsep}
  {\topsep}
  {\normalfont}
  {}
  {\bfseries\color{black}}
  {.}
  { }
  {}
\theoremstyle{remark}
\newtheorem{rem}[thm]{Remark}

\global\long\def\assign{\coloneqq}

\DeclareMathOperator*{\argmin}{argmin}

\icmltitlerunning{Wait-Less Offline Tuning and Re-solving for Online Decision Making}

\begin{document}

\twocolumn[
\icmltitle{Wait-Less Offline Tuning and Re-solving for Online Decision Making}

\begin{icmlauthorlist}
\icmlauthor{Jingruo Sun}{1}
\icmlauthor{Wenzhi Gao}{2}
\icmlauthor{Ellen Vitercik}{1,3}
\icmlauthor{Yinyu Ye}{1,4,5}
\end{icmlauthorlist}

\icmlaffiliation{1}{Department of Management Science \& Engineering, Stanford University}
\icmlaffiliation{2}{Institute for Computational and Mathematical Engineering, Stanford University}
\icmlaffiliation{3}{Department of Computer Science, Stanford University}
\icmlaffiliation{4}{Chinese University of Hong Kong (Shen Zhen)}
\icmlaffiliation{5}{Hong Kong University of Science and Technology}

\icmlcorrespondingauthor{Jingruo Sun}{jingruo@stanford.edu}

\icmlkeywords{Online Linear Programming, Sequential Decision-Making, Machine Learning}
\vskip 0.3in
]

\printAffiliationsAndNotice{}  

\begin{abstract}
    Online linear programming (OLP) has found broad applications in revenue management and resource allocation. State-of-the-art OLP algorithms achieve low regret by repeatedly solving linear programming (LP) subproblems that incorporate updated resource information. However, LP-based methods are computationally expensive and often inefficient for large-scale applications. By contrast, recent first-order OLP algorithms are more computationally efficient but typically suffer from weaker regret guarantees. To address these shortcomings, we propose a new algorithm that combines the strengths of LP-based and first-order OLP algorithms. Our algorithm re-solves the LP subproblems periodically at a predefined frequency $f$ and uses the latest dual prices to guide online decision-making. In parallel, a first-order method runs during each interval between LP re-solves and smooths resource consumption. Our algorithm achieves $\mathcal{O}(\log (T/f) + \sqrt{f})$ regret and delivers a ``wait-less'' online decision-making process that balances computational efficiency and regret guarantees. Extensive experiments demonstrate at least $10$-fold improvements in regret over first-order methods and $100$-fold improvements in runtime over LP-based methods. 
\end{abstract}

\section{Introduction}
\label{intro}

Sequential decision-making has garnered significant attention for its utility in guiding optimal strategies in dynamic environments. 
The goal is to identify effective decisions and policies in environments where knowledge of the system continuously accumulates and evolves. 
Online Linear Programming (OLP) \cite{agrawal2014dynamic} offers a powerful framework that encapsulates the core principles of sequential decision-making 
and has been extensively applied to different domains, including resource allocation \cite{bestofmanyworlds}, online advertising \cite{onlineadvertising}, and inventory management \cite{talluri2004theory}. 

We study an OLP problem where customers arrive sequentially, each requesting a combination of resources and offering a bidding price. The objective is to determine which resource requests to fulfill to maximize revenue while respecting resource constraints.
The challenge is that decisions must be made immediately and irrevocably, relying solely on historical data without knowledge of future arrivals. The goal is to minimize regret with respect to the optimal hindsight linear programming (LP) solution. 

To guide real-time decision-making, state-of-the-art OLP algorithms estimate optimal \emph{dual prices} and use them to make decisions. These algorithms fall into two main categories: LP-based and first-order methods. Specifically, LP-based methods update dual prices by repeatedly solving linear programs at each time step with all available information so far. However, the substantial computational demands limit their application in time-sensitive settings. First-order methods offer quick, incremental updates to dual prices using gradient information, but generally fall short of achieving the strong regret guarantee of LP-based methods. These trade-offs motivate an open question at the intersection of online learning and decision-making: 
\vspace{2 pt}

\noindent\textit{Can we simultaneously achieve}%
\vspace{-0.5em}

\begin{center}
\textit{low regret and computational efficiency?}
\end{center}

\vspace{-6 pt}

\paragraph{Our contributions.}
We answer this question in the affirmative. We summarize our contributions as follows: 
\begin{itemize}[leftmargin=10pt]
    \item \textit{{Parallel Multi-Stage Framework}:}
    We separate online learning and decision-making into distinct processes that interact via a feedback loop. 
    We re-solve the LP subproblems periodically at a fixed frequency and feed the updated dual price to guide decision-making until the next time that the LP is re-solved. 
    To further enhance efficiency, we apply a first-order method during the initial and final stages, restarting with the most recent learning outcomes. By integrating the LP-based and first-order techniques, our approach leverages their respective strengths and achieves a balance between regret performance and computational costs. 
    \item \textit{Regret Analysis:} 
    We unify the analysis of LP-based and first-order methods by deriving a new performance metric to account for their inter-dependency. 
    Our analysis yields a ``spectrum theorem'' that bounds regret for any feasible choice of the re-solving frequency: 
    \begin{thm}[Informal version of Theorem \ref{thm:spectrum}]
        If we re-solve the LPs every $f$ time steps within horizon $T$ and apply a first-order method in the initial and final $f$ steps, we achieve a worst-case regret of $\mathcal{O} \big(\log (T/f) + \sqrt{f} \big).$
    \end{thm}
    In particular, $f=1$ yields a pure LP-based method with $\mathcal{O}(\log T)$ regret, while $f=T$ reduces to a pure first-order method with $\mathcal{O}(\sqrt{T})$ regret. 
    By choosing an ``optimal'' $f$, one can achieve the best possible regret based on available computational resources and enable a ``wait-less'' decision-making system across all time steps. 

    \item \textit{Experiments:} 
    Through experiments across diverse distributions, we demonstrate that our algorithm achieves over a $10$-fold improvement in regret compared to first-order methods and over a $100$-fold improvement in runtime with comparable regret to LP-based methods. 
\end{itemize}

\paragraph{Key Challenges.}
OLP methods face a fundamental challenge in balancing efficient real-time decision-making and accurate dual-price learning. 
Specifically, LP-based methods ensure high-quality decisions with $\mathcal{O}(\log T)$ regret, but their computational costs increase quadratically with problem size. A natural alternative---batching customers and solving the LP once every $f$ arrivals---reduces this cost, but leaves customers waiting until the batch concludes. In contrast, first-order methods update dual prices via gradient information, allowing for faster computation. However, these methods require small step sizes to maintain decision quality, which slows their adaptation to new data. Even using different step sizes for learning and decision-making only achieves $\mathcal{O}(T^{1/3})$ regret when the distribution of customers has continuous support.
Our parallel framework addresses these issues: an LP-based method periodically refines dual prices, while a first-order method immediately processes arriving customers using the most recent dual updates, thereby eliminating delays.

 A second challenge emerges when analyzing the regret of this hybrid approach. LP-based methods rely on a stopping-time analysis (halting when resources are depleted), whereas first-order methods track constraint violations as part of the regret. Since these two formulations are fundamentally different, existing regret bounds cannot be naively combined. We overcome this challenge by introducing a unified performance metric that decomposes regret into three components: dual convergence, constraint violation, and leftover resources. This approach yields the first integrated analysis of LP-based and first-order OLP, culminating in our $\mathcal{O}(\log(T/f) + \sqrt{f})$ regret bound.


\subsection{Related Literature}

Online resource allocation and OLP problems have been widely studied under two predominant models: the stochastic input \cite{goel2008online, devanur2019} and stochastic permutation models \cite{agrawal2014dynamic, gupta2015}. 
We study the former, where each customer's resource request and bidding price are drawn i.i.d. from an unknown distribution. We use expected regret and constraint violation as the performance metric. We summarize some recent works in \cref{table:1}. 

\paragraph{LP-based Methods.} These methods solve the OLP dual repeatedly to update dual prices and make decisions. 
Early approaches enforced a fixed average resource constraint \cite{agrawal2014dynamic}, whereas recent work dynamically tracks remaining resources and enables LP-based methods to achieve $\mathcal{O}(\log T)$ regret under continuous support and non-degeneracy assumptions \cite{li2022online}. Variants include multi-secretary problems \cite{bray2019logarithmic}, regularized resource constraints \cite{ma2024optimal}, and finite-support distributions yielding constant regret \cite{chen2024improved}. 

\paragraph{First-order Methods.} These methods generate decisions using gradient updates without solving LPs, enabling efficient computation. They achieve $\mathcal{O}(\sqrt{T})$ regret with mirror descent \cite{sun2020, bestofmanyworlds} and $\mathcal{O}(T^{3/8})$ under finite support and non-degeneracy assumptions \cite{sun2020near}. Variants include proximal updates \cite{gao2023solving}, momentum-based mirror descent \cite{balseiro2022online}, resource adjustments \cite{ma2024optimal}, and restart strategies yielding $\mathcal{O}(T^{1/3})$ regret \cite{gao2024decoupling}. 

\paragraph{Delay in Decision-making.} 

\begin{table*}[t]
    \centering
        \caption{Performances of Dual Algorithms in Recent OLP Literature
        \label{table:1}}
        \resizebox{\textwidth}{!}{%
    \begin{tabular}{clclc}
        \toprule
        Paper & Setting & Algorithm & Regret & Decision-Making \\
        \midrule
        {\cite{li2022online}} & Bounded, continuous support, non-degeneracy & LP-based & $\mathcal{O}(\log T \log \log T)$ & Delay \\
        {\cite{bray2019logarithmic}} & Bounded, continuous support, non-degeneracy & LP-based & $\mathcal{O} (\log T)$ & Delay \\
        \cite{chen2024improved} & Bounded, finite support, non-degeneracy &  LP-based & $\mathcal{O}(1)$ & Delay \\
        \cite{infrequent2024} & Bounded, finite support &  LP-based & $\mathcal{O}(1)$ & Delay \\
        {\cite{xu2024online}} & Bounded, continuous support, non-degeneracy & LP-based &
        $\mathcal{O} (\log (T/f))$ & Delay \\
        \midrule
        \textbf{This paper} & \textbf{Bounded, continuous support, non-degeneracy} & \textbf{LP-based \& First-order} & $\textcolor{red}{\mathcal{O}(\log (T/f) + \sqrt{f})}$ & \textbf{No Delay} \\
        \midrule
        {\cite{sun2020}} & Bounded & First-order &
        $\mathcal{O} (\sqrt{T})$ & No Delay \\
        {\cite{bestofmanyworlds}} & Bounded & First-order & $\mathcal{O} (\sqrt{T} )$ & No Delay \\ 
        {\cite{gao2023solving}} & Bounded & First-order & $\mathcal{O}(\sqrt{T})$ & No Delay \\
        {\cite{sun2020near}} & Bounded, finite support, non-degeneracy & First-order & $\mathcal{O}(T^{3/8})$ & No Delay \\
        {\cite{gao2024decoupling}} & Bounded, continuous support, non-degeneracy & First-order & $\mathcal{O} (T^{1/3})$ & No Delay \\
        \bottomrule
    \end{tabular}
    }
\end{table*}

Delays arise from the time-consuming process of solving the large-scale, up-to-date LP subproblems for each customer. \citet{golrezaei2023} study a mix of impatient and partially patient customers, while \citet{xie2023benefits} show that batching requests can reduce regret. Concurrent work \cite{xu2024online} reduces delay by solving LPs in batches but assumes a lower bound on resource requests and still requires waiting in initial and final batches. To achieve delay-free decisions, we shift the re-solving process offline and only fine-tune the solution online. Compared with previous works, our framework imposes standard assumptions, achieves $\mathcal{O}(\log (T/f) + \sqrt{f})$ regret, and ensures ``wait-less'' decision-making throughout the entire horizon. 

\paragraph{Paper organization.} The rest of the paper is organized as follows. \cref{problem_setup} introduces the problem formulation and assumptions. Section \ref{algorithm} proposes our algorithms and main theoretical guarantee: a $\mathcal{O}(\log(T/f)+\sqrt{f})$ regret bound.
Section \ref{experiments} presents experiments to validate our theory. 

\section{Problem Setup} \label{problem_setup}

We use $\| \cdot \|$ to denote the Euclidean norm and $\langle \cdot \rangle$ to denote Euclidean inner product. 
Bold letters $\bm{A}$ and $\bm{a}$ denote matrices and vectors, respectively. 
Subscript $(\cdot)_{it}$ denotes the index for resource type $i$ at time $t$.
The notation $(\cdot)^+ = \max \{ \cdot, 0 \}$ denotes the element-wise positive part function, and $\mathbb{I} ( \cdot )$ denotes the 0-1 indicator function. 

\subsection{OLP Formulation}
We study an online resource allocation problem over the time horizon $T$ under a \emph{stochastic input model}. Initially, we have an inventory vector $\bm{b} \in \mathbb{R}^m$, representing $m$ resource types. At each time step $t$, a customer arrives with a request sampled i.i.d. as $(r_t, \bm{a}_t) \sim \mathcal{P}$, where $\bm{r} = (r_1, \ldots, r_T)^\top \in \mathbb{R}^T$ is the offered payment (bid), $\bm{A} = (\bm{a}_1, \ldots, \bm{a}_T) \in \mathbb{R}^{m \times T}$ is the matrix of customers' resource demands, and $\mathcal{P}$ is a fixed, unknown distribution. We must decide whether to accept or reject each request, represented by the decision variables $\bm{x} = (x_1, \ldots, x_T) \in \{0, 1\}^{T}$. The goal is to maximize the cumulative reward. This problem can be formulated as the following OLP, referred to as the \emph{primal linear program (PLP)}: 
\begin{align} \tag{PLP} \label{olp:primal}
    \max_{\bm{0} \leq \bm{x} \leq \bm{1}} \langle \bm{r}, \bm{x} \rangle 
    \quad \text{s.t.} \quad \bm{Ax} \leq \bm{b}. 
\end{align}
The dual problem of \eqref{olp:primal} is given by
\begin{align} \tag{DLP} \label{olp:dual}
    \min_{(\bm{p}, \bm{y}) \geq \bm{0}} \langle \bm{b}, \bm{p} \rangle + \langle \bm{1}, \bm{y} \rangle 
    \quad \text{s.t.} \quad \bm{A}^\top \bm{p} + \bm{y} \geq \bm{r}
\end{align}
where $\bm{p}$ is the vector of \emph{dual prices.}
Let $\bm{d} = \bm{b} / T \in \mathbb{R}^m$ denote the initial average resource. As demonstrated by
\cite{sun2020}, \eqref{olp:dual} can be written as 
\begin{align} \label{opt:real_dual}
    \min_{\bm{p} \geq \bm{0}} f_T(\bm{p}) \assign \tfrac{1}{T} \textstyle\sum_{t=1}^T \left[ \langle \bm{d}, \bm{p} \rangle + (r_t - \langle \bm{a}_t, \bm{p} \rangle )^+ \right]
\end{align}

This formulation can be written as a $T$-sample approximation of the following stochastic program:
\begin{align} \label{opt:e_dual}
    \min_{\bm{p} \geq \bm{0}} f(\bm{p}) \assign \mathbb{E}[f_T(\bm{p})] = \bm{d}^\top \bm{p} + \mathbb{E} [(r - \bm{a^\top p})^+]
\end{align}
where the expectation is taken with respect to $(r, \bm{a})\sim \mathcal{P}$. 

We define optimal solutions to the $T$-sample approximation problem \eqref{opt:real_dual} and stochastic program \eqref{opt:e_dual} respectively as 
\begin{align*}
    \bm{p}_T^* = \arg \min_{\bm{p} \geq \bm{0}} ~f_T (\bm{p})
    \quad \text{and}\quad
    \bm{p}^* = \arg \min_{\bm{p} \geq \bm{0}} ~f (\bm{p}).
\end{align*}
The decision variable $x_t^*$ can be established from the complementary slackness condition as
\begin{align} \label{eqn:kkt}
    x_t^* = \left\{
    \begin{array}{ll}
      \ 0, & r_t < \bm{a}_t^\top \bm{p}_T^*, \vspace{2 pt} \\
      \ 1, & r_t > \bm{a}_t^\top \bm{p}_T^*
    \end{array} \right.
\end{align}
and $x_t^* \in [0, 1]$ if $r_t = \bm{a}_t^\top \bm{p}_T^*$. 

This connection between primal and dual solutions inspires OLP algorithms that make decisions based on dual prices: 
\begin{align} \label{decision_rule}
    x_t = \mathbb{I}(r_t \geq \bm{a}_t^\top \bm{p}_t).
\end{align}

\subsection{Algorithms for OLP}

We summarize two main dual-based OLP algorithms. 

\paragraph{LP-based Method.}\label{sec:LP}

The LP-based method \citep{li2022online} calculates
dual prices by re-solving the online LP problem at each time step. Specifically, define $d_{it} = b_{it}/(T-t)$ as the average remaining resource for type $i$ at time $t$. The resulting optimization problem can be viewed as a $t$-sample approximation to the stochastic program specified by $\bm{d}_t = (d_{it}, ..., d_{mt})^\top$ as 
\begin{align}\label{opt:d_dual}
    \min_{\bm{p} \geq \bm{0}} & ~~~~ f_{\bm{d}_t}(\bm{p}) \assign \bm{d}_t^\top \bm{p} + \mathbb{E} [(r - \bm{a^\top p})^+]
\end{align}
By updating $\bm{d}_t$, this method incorporates past decisions when computing $\bm{p}_t$. If resources are over-utilized in earlier periods, the supply decreases, prompting us to raise the dual price and be more selective with future orders. We outline this method in \cref{alg:lp-based} of \cref{app-lp}. 


\paragraph{First-order Method.}\label{sec:subgradient}

The first-order method \citep{sun2020} calculates dual prices via online subgradient updates. We maintain a static average resource $\bm{d} = \bm{b} / T$, compute $x_t$ as per \eqref{decision_rule}, and update the dual price as 
\begin{align*}
    \bm{p}_{t+1} &= (\bm{p}_t - \alpha_t (\bm{d} - \bm{a}_t x_t))^+
\end{align*}
where the subgradient term evaluated at $\bm{p}_t$ is \begin{align*}
    \bm{d} - \bm{a}_t x_t \in \partial_{\bm{p} = \bm{p}_t} \big( \bm{d}^\top \bm{p} + (r_t - \bm{a}_t^\top \bm{p})^+ \big).
\end{align*}
This process can be interpreted as a projected stochastic subgradient method for solving \eqref{opt:real_dual}. It reduces computational cost by requiring only a single pass through the data and eliminates solving LPs explicitly. A restart strategy improves the first-order method to have $\mathcal{O}(T^{1/3})$ regret \cite{gao2024decoupling}. \cref{alg:first-order} and \cref{alg:first-order_gao} in \cref{app_fo} summarizes this method.

\subsection{Performance Metrics}

We aim to design algorithms that optimize a bi-objective performance measure involving regret and constraint violation. The regret measures the difference between the objective value of the algorithm's output and that of the true optimal solution, while the constraint violation measures the degree to which the algorithm's output fails to meet the given constraints. We denote the offline optimal solution to \eqref{olp:primal} by $\bm{x}^* = (x_1^*, \ldots, x_T^*)$, and the online algorithm output by $\bm{x} = (x_1, \ldots, x_T)$. Then, we define the regret $r(\bm{x})$ and resource violation $v(\bm{x})$ as
\begin{align}
    r(\bm{x}) &\assign \langle \bm{r}, \bm{x}^* \rangle - \langle \bm{r}, \bm{x} \rangle, \label{def:regret} \\
    v(\bm{x}) &\assign \lVert (\bm{Ax} - \bm{b})^+ \rVert. \label{def:vio}
\end{align}

Therefore, we define the following bi-objectives for evaluating an algorithm's worst-case performance:
\begin{align} \label{def:bi-obj}
    \Delta_T = \sup_{\mathcal{P} \in \Xi} \mathbb{E}_{\mathcal{P}}[r(\bm{x}) + v(\bm{x})]
\end{align}
where $\Xi$ denotes a family of distributions satisfying regularity assumptions specified later. 

This metric is commonly used for first-order OLP algorithms \cite{gao2023solving, sun2020} and is also aligned with the literature on online convex optimization with constraints \cite{JMLR:v13:mahdavi12a, yu2017onlineconvexopt}.
Integrating $r(\bm{x})$ and $v(\bm{x})$ into a single performance measure promotes balanced resource consumption over the decision horizon. 
In \cref{algorithm}, we derive a decomposition of \eqref{def:bi-obj} that unifies the regret analysis for all OLP methods. 

\subsection{Assumptions and Auxiliary Results}

We adopt the following assumptions regarding the stochastic inputs. These assumptions are standard in the online learning literature \cite{li2022online, jiang2024, xu2024online}. In particular, we require the input data to be bounded, follow a linear growth, and be non-degenerate.

\begin{ass}[Boundedness]\label{ass:1}
    We assume
    \begin{enumerate}
        \item[(a)] The order inputs $\{(r_t, \bm{a}_t) \}_{t=1}^T$ are generated i.i.d from an unknown distribution $\mathcal{P}$. 
        \item[(b)] There exist constants $\bar{r}, \bar{a} > 0$ such that $|r_t| \leq \bar{r}$ and $\lVert \bm{a}_t \rVert_{\infty} \leq \bar{a}$ almost surely for $t = 1, ..., T$. 
        \item[(c)] The average resource capacity $\bm{d} = \bm{b} / T$ satisfies $d_i \in [\underline{d}, \bar{d}]$ for some $\bar{d} > \underline{d} > 0$ for any $i = 1, ..., m$.
    \end{enumerate}
\end{ass}

In this assumption, (a) states that $\{(r_t, \bm{a}_t) \}_{t=1}^T$ are independent of each other, but we allow dependencies between their components. Part (b) introduces the bounds $\bar{r}, \bar{a}$ solely for analytical purposes. This is a minimal requirement on $(r_t, \bm{a}_t)$ compared to previous work \cite{agrawal2014dynamic, li2022online, xu2024online}. Part (c) requires the average resource to grow linearly with $T$, ensuring that a constant fraction of $x_t$ values can be set to 1. Consequently, the number of fulfillable orders is proportional to $T$, facilitating a stable service level over time.

\begin{ass}[Uniform Non-degeneracy]\label{ass:2}
    We assume
    \begin{enumerate}
        \item[(a)] The second-moment matrix $\mathbb{E}[\bm{a} \bm{a}^\top]$ is positive definite with minimum eigenvalue $\lambda$.
        \item[(b)] There are constants $\mu, \nu$ such that for any $(r, \bm{a}) \sim \mathcal{P}$, 
        \begin{align*}
            & \nu |\bm{a}^\top (\bm{p} - \bm{p}^*)| \\
            \leq \ & \left|\mathbb{P}(r \geq \bm{a}^\top \bm{p} \mid \bm{a}) - \mathbb{P}(r \geq \bm{a}^\top \bm{p}^* \mid \bm{a})\right| \\
            \leq \ & \mu |\bm{a}^\top (\bm{p} - \bm{p}^*)|
        \end{align*}
        holds for all \[\bm{p} \in \mathcal{V}_{\bm{p}} \assign \left\{\bm{p} \in \mathbb{R}^m: \bm{p} \geq \bm{0}, \lVert \bm{p} \rVert \leq \frac{\bar{r}}{\underline{d}} \right\}\] and $\bm{d} \in \mathcal{V}_{\bm{d}} \assign [\underline{d}, \bar{d}]^m$. 
        \item[(c)] The optimal $\bm{p}^*$ satisfies $p_i^* = 0$ if and only if $d_i - \mathbb{E}[a_i \mathbb{I}(r > \bm{a}^\top \bm{p}^*)] > 0$ for all $\bm{d} \in \mathcal{V}_{\bm{d}}$ and $i \in [m]$. 
    \end{enumerate}
\end{ass}

In this assumption, part (b) ensures that the cumulative distribution of the reward given the resource consumption request $r | \bm{a}$ is continuous and exhibits a stable growth rate. Part (c) requires strict complementarity for the optimal solutions of the stochastic program in \eqref{opt:e_dual}, which is a non-degeneracy condition for both the primal and dual LPs. 

According to stochastic program \eqref{opt:e_dual}, we define the binding and non-binding index sets as: 
\begin{align}\label{eqn:binding}
    I_B &= \{i: d_i - \mathbb{E}[a_i \mathbb{I}(r > \bm{a}^\top \bm{p}^*)] = 0\}, \nonumber \\
    I_N &= \{i: d_i - \mathbb{E}[a_i \mathbb{I}(r > \bm{a}^\top \bm{p}^*)] > 0\}.
\end{align}
By \cref{ass:2}(c), these sets are complements, as $I_B \cap I_N = \emptyset$ and $I_B \cup I_N = \{1, ..., m\}$. 

\section{Parallel Multi-Phase OLP Algorithm} \label{algorithm}

We now present our algorithm for parallel multi-phase online learning and decision-making. Our approach is motivated by the challenges of existing methods. Specifically, the LP-based method has a strong worst-case regret bound of $\mathcal{O}(\log T)$, but its high computational costs lead to decision-making delays. Meanwhile, the first-order method updates decisions efficiently but suffers a high regret of $\mathcal{O}(\sqrt{T})$. By combining the strengths of these two methods, our new framework balances decision-making quality and computational efficiency. 

\subsection{Algorithm Design}

\begin{figure*}[t]
    \centering
    \vspace{-3 cm}
    \includegraphics[scale=0.54]{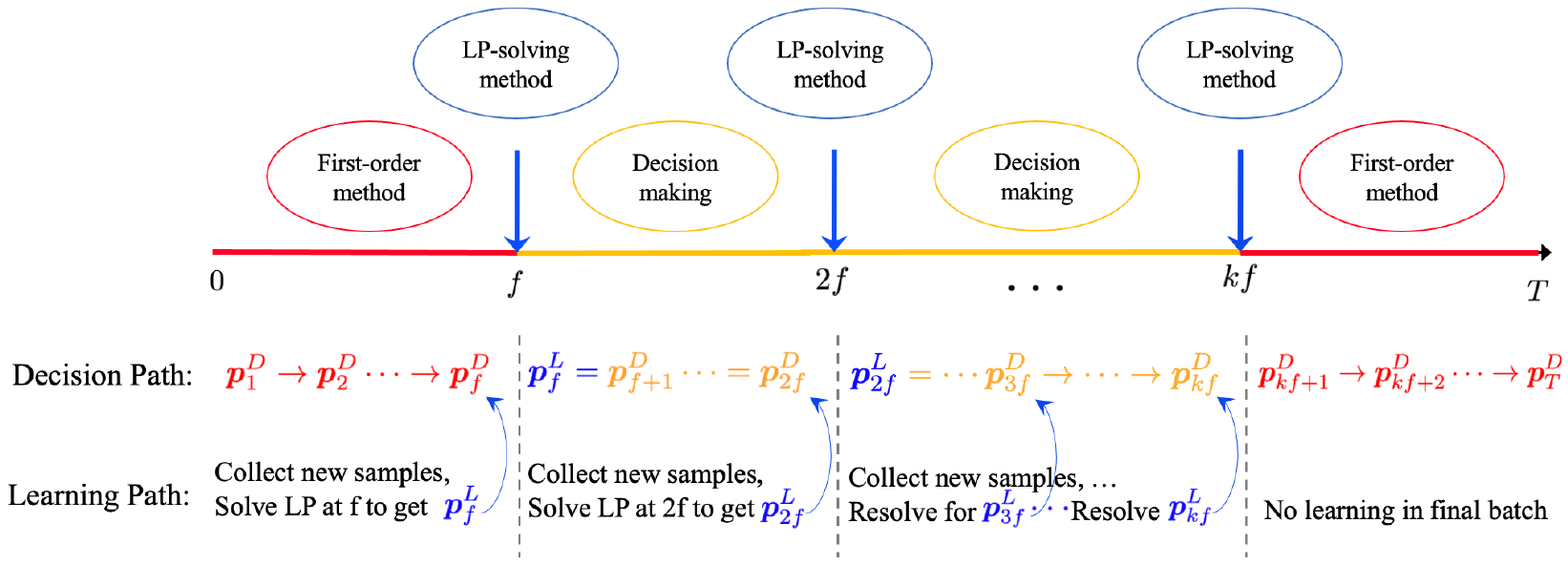}
    \vspace{-3.1 cm}
    \caption{Algorithm \ref{alg:main-1} illustration of parallel paths and the interactions between online learning and decision-making. Decisions are generated based on 1) the LP-based method (blue) with frequency $f$, 2) the first-order method (red) for the initial and final phases (with a warm start), and 3) employing the latest dual price (yellow) during intermediate phases.
    \label{fig:alg1}}
    \vskip -0.1in
\end{figure*}

We establish our framework with the following design:
\begin{enumerate}[leftmargin=10pt]
    \item We maintain two parallel paths of online learning and online decision-making. The online learning results are periodically sent to the decision-making path as a re-start point to guide subsequent updates.
    \item In the online learning path, we employ a streamlined LP-based method, which only re-solves the updated OLP problem according to a predefined frequency, reducing the computational overhead.
    \item In the online decision-making path, we apply the first-order method during the initial and final batches. The learning rate is optimally tuned for these two intervals.
\end{enumerate}


Figure \ref{fig:alg1} illustrates our framework, which generates two parallel sequences of dual prices: $\{\bm{p}_t^D\}_{t=1}^T$ from the first-order method and $\{\bm{p}_t^L\}_{t=1}^T$ from the LP-based method. The algorithm proceeds in batches of length $f$. In the first batch, it uses the first-order method to iteratively update the dual prices, thereby guiding decision-making. At the end of this batch, it applies the LP-based method to obtain a refined dual price and passes it to the decision-making path. The LP re-solving occurs only once per batch, making the approach computationally efficient. Formally, we re-solve the OLP at every time $t$ satisfying $t \leq kf$ and $t \bmod f = 0,$ where $k =\lfloor T/f\rfloor$ is the number of batches. 

During intermediate batches, decisions are made based on the most recent dual prices computed from the LP at the end of the previous batch. 
The algorithm restarts the subgradient updates only in the final batch to guide the remaining decisions. 
Algorithm \ref{alg:main-1} summarizes this approach. 

\begin{algorithm}[t]
\caption{Parallel Multi-Phase OLP Algorithm
\label{alg:main-1}}
    \KwIn{total resource $\bm{b}$, time horizon $T$, average resource $\bm{d} = \bm{b} / T$, initial dual price $\bm{p}_1 = \bm{0}$, re-solving frequency $f$, and number of batches $k = \lfloor T / f \rfloor$}
    \For{$t = $ \rm{$1$ to $T$ }}{

    Observe $(r_t, \bm{a}_t)$ and make decision $x_t$ as rule \eqref{decision_rule}

    Update constraint for $i = 1, ..., m$: 

    \hspace{15 pt} remaining resource $b_{it} = b_{i, t-1} - a_{it} x_t$

    \hspace{15 pt} average remaining resource $d_{it} = \frac{b_{it}}{T-t}$

    \If{$t \leq f$ \rm{or} $t \geq kf$}{
    \rm{Update learning rate} $\alpha_t = \begin{cases}
        \mathcal{O}(1/f^{1/2}) & t \leq f \\
        \mathcal{O}(1/f^{2/3}) & t \geq kf
    \end{cases}$
    
    \rm{Compute subgradient and update dual price $\bm{p}_{t+1}$:}
    \vspace{-3 pt}
    \begin{align*} 
        \bm{p}_{t+1} &= (\bm{p}_t - \alpha_t (\bm{d} - \bm{a}_t x_t))^+
    \end{align*}
    \vspace{-15 pt}
    }

    \If{$t \bmod f = 0$ \rm{and} constraints are not violated}{
    \rm{Solve OLP and update dual price $\bm{p}_{t+1}$:}
    \vspace{-3 pt}
    \begin{align*} 
        \bm{p}_{t+1} = \argmin_{\bm{p} \geq \bm{0}} \ \bm{d}_t^\top \bm{p} + \tfrac{1}{t} \textstyle\sum_{j=1}^t (r_j - \bm{a}_j^\top \bm{p})^+
    \end{align*}
    \vspace{-10 pt}
    }

    \Else{
    \rm{Update dual price $\bm{p}_{t+1}$ to be the most recent solution $\bm{p}_{t+1} = \bm{p}_t$.}
    }
    }
\end{algorithm}

Algorithm \ref{alg:main-1} balances online learning and efficient decision-making. It is responsive to dynamic environments, adapts to the latest information, and reduces computational cost by periodic re-solving. We thus establish a ``wait-less'' online decision-making framework where each customer's order is processed immediately without delays from earlier requests or large-scale LP computations. 

\subsection{Algorithm Analysis}

We decompose the performance metric of Algorithm~\ref{alg:main-1} into three key components.
All proofs (including essential properties of the dual price $\bm{p}_t$) are in Appendix \ref{app:framework} and \ref{app:main}. 

\begin{thm} [Performance Metric] \label{thm:regret}
    Under Assumptions \ref{ass:1} and \ref{ass:2}, the performance $\Delta_T$ of Algorithm \ref{alg:main-1} satisfies
    \begin{align}
        \Delta_T \ & \leq \ \mu \bar{a}^2 \textstyle\sum_{t=1}^T \mathbb{E} \left[ \lVert \bm{p}_t - \bm{p}^* \rVert^2 \right] + \mathbb{E} \left[ \lVert (\bm{Ax} - \bm{b})^+ \rVert \right] \nonumber \\
        & \quad + \lVert \bm{p}^* \rVert \cdot \mathbb{E} \left[ \lVert (\bm{b} - \bm{Ax})^{B+} \rVert \right].
    \end{align}
    where $(\cdot)^{B+}$ indicates the projection of binding terms onto the positive orthant. 
\end{thm}

Based on the definition of $\Delta_T$ \eqref{def:bi-obj}, we derive the performance metric as the sum of three key components: dual convergence, remaining resources, and binding constraint violation.
This bound demonstrates that our algorithm encourages 1) smooth and balanced resource utilization, and 2) full resource consumption by the end of the time horizon. 
Building on Theorem~\ref{thm:regret}, we provide a ``spectrum theorem'' for the algorithm's performance. 

\begin{thm}\label{thm:spectrum}
     Under Assumptions \ref{ass:1} and \ref{ass:2}, the worst-case performance $\Delta_T$ of Algorithm \ref{alg:main-1} is bounded by
     \begin{align}
        \Delta_T \in  \mathcal{O} \big(\log \left( \tfrac{T}{f} \right) + \sqrt{f} \big)
     \end{align}
     where $T / f$ represents the number of re-solving batches and $f$ is the length of each batch.
 \end{thm}

\begin{rem}[Spectrum Theorem]
    We elucidate the trade-offs in total regret induced by using LP-based methods with $\mathcal{O}(\log T)$ regret and first-order methods with $\mathcal{O}(\sqrt{T})$ regret. 
    Our algorithm introduces a re-solving frequency $f \in [1, T]$ and achieves $\mathcal{O}(\log(T/f) + \sqrt{f})$ regret, which recovers previous results with extreme cases of $f = 1$ and $f = T$.
    Specifically, the first-order method gives us the regret of $\sqrt{f}$ for the first batch and $f^{1/3}$ for the last batch, and the LP-based method contributes the regret of $\log(T/f)$. 
\end{rem}

\begin{rem}[Warm Start]
    We can obtain a tighter regret bound of 
    $\mathcal{O}(\log(T/f) + f^{1/3})$
    given a warm start of the initial dual price satisfying $\lVert \bm{p}_0 - \bm{p}^* \rVert \leq f^{-1/3}$ or if we use the LP-based method at each time step for the first batch. 
\end{rem}

\begin{rem}[Learning Rate Selection]
    We select the best learning rate to minimize the regret upper bound consisting of \eqref{def:regret} and \eqref{def:vio}. For the first batch, the regret grows linearly with $\alpha_t$ while the constraint violation is inversely proportional to $\alpha_t$. To achieve the tightest bound, we balance this trade-off by selecting the learning rate that minimizes the overall expression, which yields the optimal choice $\alpha_t = 1 / \sqrt{f}$ as derived in Theorem \ref{thm:fo-regret-1}. A similar analysis for the final batch leads to the choice $\alpha_t = 1/f^{2/3}$ as shown in Theorem \ref{thm:fo-regret-2}. 
\end{rem}

\paragraph{Technical Intuitions.}
Theorem \ref{thm:regret} decomposes the performance metrics to make the regret analysis compatible between the LP-based and first-order methods.
\vspace{-3 pt}
\begin{itemize}[leftmargin=10pt]
    \item \textbf{LP-based Method:} As Theorem \ref{thm:regret} suggests, achieving small regret requires $\bm{Ax}-\bm{b}$ to be close to zero. To enforce this, we impose a stronger condition to track the average remaining resource---if $\bm{d}_t$ exceeds the allowed limit, we manually set the dual prices to zero to accept all subsequent orders. This approach enables us to eliminate the ``stopping time'' argument and make the analysis compatible with our new framework.
    \item \textbf{First-order Method:} The regret improvement comes from the final batch since it starts with the LP-derived learning result $\bm{p}_{kf}$, which lies within a $\mathcal{O}(1/\sqrt{kf})$-sized neighborhood around $\bm{p}^*$. Using the subgradient method, we bound the three components in \cref{thm:regret} in terms of $\lVert \bm{p}_{t} - \bm{p}^* \rVert$ and express them in terms of $f, k, T,$ and $\alpha_t$. Optimizing the learning rate to $\alpha_t = f^{-2/3}$ yields an improved regret of $\mathcal{O}(f^{1/3})$ in the final batch.
\end{itemize}


\subsection{Algorithm Extension}

\begin{figure*}[t]
    \centering
    \vspace{-3 cm}
    \includegraphics[scale=0.54]{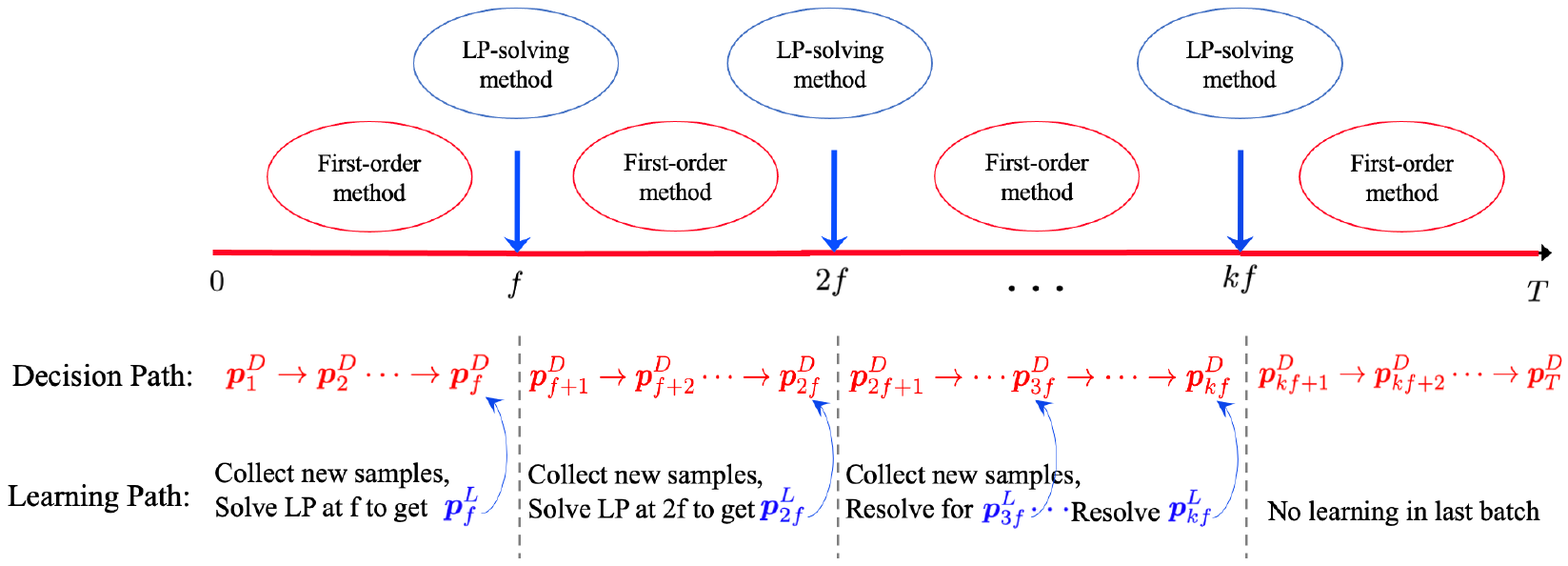}
    \vspace{-3.2 cm}
    \caption{Algorithm \ref{alg:main-2} illustration of parallel paths with multi-time restart. Decisions are generated based on 1) the LP-based method (blue) with frequency $f$ and 2) the first-order method (red) during re-solving intervals with a warm start. 
    \label{fig-alg-2}}
\end{figure*}

We propose an enhanced version of Algorithm \ref{alg:main-1} that employs the first-order method for decision-making between two consecutive LP resolves. 
Rather than making decisions solely with the most recently solved dual price from the learning path, Algorithm \ref{alg:main-2} treats it as a new starting point for each re-solving interval and adopts a smaller step size for subgradient updates to avoid deviating too far from the LP-guided solutions. Figure \ref{fig-alg-2} illustrates this framework. 
Specifically, for each interval $[jf, (j+1)f]$, we attain the new start $\bm{p}_{jf}$ from the learning path at time $jf$ and continue to fine-tune the dual price $\{ \bm{p}_t \}_{t=jf}^{(j+1)f}$ using the first-order method with a learning rate $\alpha_t = \mathcal{O}(1/t)$. Algorithm \ref{alg:main-2} summarizes this approach. 

As we illustrate in our experiments (Section~\ref{experiments}), the incorporation of an intermediate first-order method improves the algorithm's stability and ensures smooth resource consumption. Therefore, the multi-restart mechanism results in better performance during the final batch and improves the algorithm's total performance. 

\begin{algorithm}[t]
\caption{Enhanced Multi-Start OLP Algorithm
\label{alg:main-2}}
    \KwIn{total resource $\bm{b}$, horizon $T$, average resource $\bm{d} = \bm{b} / T$, initial dual $\bm{p}_1 = \bm{0}$, re-solving frequency $f$}
    
    \For{$t = $ \rm{$1$ to $T$ }}{

    Observe $(r_t, \bm{a}_t)$ and make decision $x_t$ as rule \eqref{decision_rule}

    Update constraint for $i = 1, ..., m$: 

    \hspace{15 pt} remaining resource constraint $b_{it} = b_{i, t-1} - a_{it} x_t$

    \hspace{15 pt} average resource capacity $d_{it} = \frac{b_{it}}{T-t}$

    \If{$t \bmod f \neq 0$}{
    \rm{Update learning rate $\alpha_t = \mathcal{O}(1/t)$}
    
    \rm{Compute subgradient and update dual price $\bm{p}_{t+1}$:}
    \vspace{-3 pt}
    \begin{align*}
        \bm{p}_{t+1} &= (\bm{p}_t - \alpha_t (\bm{d} - \bm{a}_t x_t))^+
    \end{align*}
    \vspace{-15 pt}
    }

    \If{$t \bmod f = 0$ \rm{and} constraints are not violated}{
    \rm{Solve OLP and update dual price $\bm{p}_{t+1}$:}
    \vspace{-3 pt}
    \begin{align*}
        \bm{p}_{t+1} = \argmin_{\bm{p} \geq \bm{0}} \ \bm{d}_t^\top \bm{p} + \tfrac{1}{t} \textstyle\sum_{j=1}^t (r_j - \bm{a}_j^\top \bm{p})^+
    \end{align*}
    \vspace{-10 pt}
    }
    }
\end{algorithm}

\subsection{Algorithm Application}

Our motivation in designing Algorithm \ref{alg:main-1} is to effectively balance computational efficiency and decision optimality. Building on the Spectrum Theorem \ref{thm:spectrum}, this section aims to translate our theoretical results into practical applications. 

We formulate a new optimization problem to find the optimal re-solving frequency that minimizes the regret in \cref{thm:spectrum}, taking into account computational resource capacities. 
Denote $c_1(\cdot), c_2(\cdot)$ as computational cost functions for LP-based and first-order methods, and $R$ as the total computational resource capacity. The optimal value of $f$ is determined by solving the following optimization problem:
\vskip -0.2in
\begin{align} 
    \min_{f \in \{1, \dots, T\}} & ~~~~ \log \Big( \tfrac{T}{f} \Big) + \sqrt{f} \nonumber \\
    \text{s.t. } \ \ \ \ &~\quad c_1(k) + 2c_2(f) \leq R. \label{opt:comp}
\end{align}
Specifically, if we use the interior-point method or the simplex method as the LP solver, the computational cost is $m^2(m + t)$ for any time $t$. The first-order method updates gradients in constant time. The following proposition provides a concrete example in practice. 

\begin{prop}[Optimal Re-solving Frequency]
\label{prop:f}
    Given a fixed computation resource capacity $R$, if we use the interior-point or simplex method as the LP solver in Algorithm \ref{alg:main-1}, we can instantiate Constraint~\eqref{opt:comp} as 
    \begin{align*}
        \textstyle\sum_{j=1}^k (m^2 (m+jf)) + 2mf \leq R.
    \end{align*}
\end{prop} \vskip -0.1in
This proposition enables users to determine the optimal re-solving frequency that balances regret and computational cost based on available computational resources. 

\section{Numerical Experiments} \label{experiments}

\begin{figure*}[t]
    \vskip -0.2in
    \centering
    \begin{minipage}{0.4\textwidth}
        \centering
        \addtocounter{table}{1}
        \captionsetup{type=table}
        \caption{Algorithms comparison.
        \label{table:4}}
        \resizebox{\textwidth}{!}{
        \begin{tabular}{crrr}
            \toprule
            $T$ & Regret & Algorithm & Compute Time (s) \\
            \midrule
            \multirow{5}{*}{$10^4$} & 115.32 & $\mathcal{O}(T^{1/2})$ First-Order & $0.008$ \\
            & 60.39 & $\mathcal{O}(T^{1/3})$ First-Order & $0.013$ \\
            & 3.50 & LP-based & $123.497$ \\
            & \textbf{9.12} & Algorithm \ref{alg:main-1} & \textbf{1.8} \\
            & \textbf{5.09} & Algorithm \ref{alg:main-2} & \textbf{1.8} \\
            \midrule
            \multirow{5}{*}{$10^5$} & 203.20 & $\mathcal{O}(T^{1/2})$ First-Order & $0.118$ \\
            & 73.07 & $\mathcal{O}(T^{1/3})$ First-Order & $0.108$ \\
            & 3.88 & LP-based & $>3600$ \\
            & \textbf{9.41} & Algorithm \ref{alg:main-1} & \textbf{56.9} \\
            & \textbf{6.36} & Algorithm \ref{alg:main-2} & \textbf{56.8} \\
            \midrule
            \multirow{5}{*}{$10^6$} & 351.91 & $\mathcal{O}(T^{1/2})$ First-Order & $1.211$ \\
            & 115.39 & $\mathcal{O}(T^{1/3})$ First-Order & $1.577$ \\
            & 5.50 & LP-based & $>100000$ \\
            & \textbf{11.65} & Algorithm \ref{alg:main-1} & \textbf{2155.9} \\
            & \textbf{7.09} & Algorithm \ref{alg:main-2} & \textbf{2242.1} \\
            \bottomrule
        \end{tabular}
        }
    \end{minipage}
    \quad \quad
    \begin{minipage}{0.43\textwidth}
        \vskip 0.25in
        \centering
        \addtocounter{figure}{1}
        \caption{Regret for various algorithms.
        \label{fig:2}}
        \vskip 0.02in
        \resizebox{\textwidth}{!}{
            \includegraphics{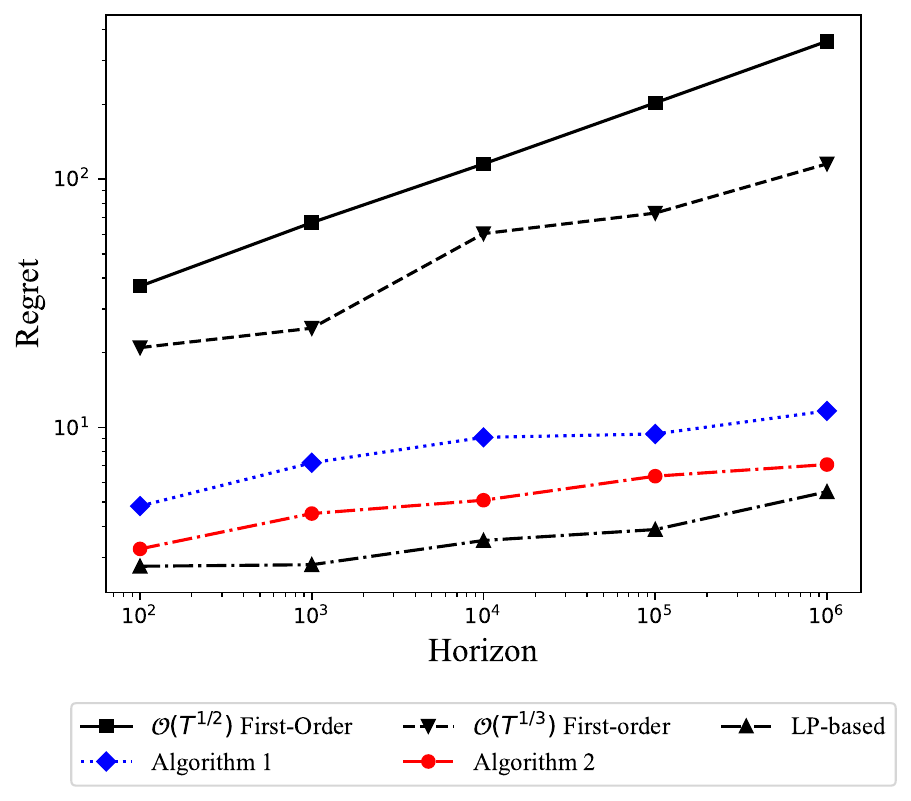}
        }
    \end{minipage}
\end{figure*}

We conduct extensive experiments to evaluate our algorithm's performance and validate our theoretical results. This section is divided into two parts. In the first part (Section \ref{exp:1}), we evaluate Algorithms \ref{alg:main-1} and \ref{alg:main-2} across different choices of re-solving frequency. 
In the second part (Section \ref{exp:2}), we compare our algorithm with LP-based and first-order methods in terms of regret and running time. All implementations can be found at \href{https://github.com/Jingruo/Wait-Less-Online-Decision-Making.git}{GitHub Link}. 

We consider the following distributions:
\begin{align*}
    & \text{Input I:} \ \ a_{it} \sim \text{Unif} [0, 2], r_t \sim \text{Unif} [0, 10] \\
    & \text{Input II:} \ \ a_{it} \sim \mathcal{N} (0.5, 1), r_t \sim \mathcal{N} (0.5m, m)
\end{align*}
Learning rates are selected as specified in Section \ref{algorithm}: 
\begin{align*}
    & \text{Algorithm 1: } \alpha_t = 
        \begin{cases}
            \mathcal{O}(1 / f^{1/2}) & t \leq f \\
            \mathcal{O}(1 / f^{2/3}) & t \geq kf 
        \end{cases} \\
    & \text{Algorithm 2: } \alpha_t = \mathcal{O}(1/t)
\end{align*}

\subsection{Regret under Varying Re-solving Frequencies} \label{exp:1}

We choose $m=1$ and generate the sequence $\{(r_t, \bm{a}_t)\}_{t=1}^T$ randomly from the uniform (Input I) and normal (Input II) distributions, and include more complex distributions in Appendix \ref{supp_exp:new}. The time horizon $T$ spans evenly over $[10^2, 10^6]$, the initial average resource is sampled as $d_i \sim \text{Uniform}[1/3, 2/3]$, and $f \in \{T^{1/3}, T^{1/2}, T^{2/3}\}$ representing high, medium, and low re-solving frequencies. 
We report the average result over 100 trials for each experiment. We use the classic first-order method with $\mathcal{O}(T^{1/2})$ regret \cite{sun2020} as a baseline. 
Results are summarized in Table \ref{table:2} and Figure \ref{fig:1} on a logarithmic scale.

We analyze the absolute value of regret and its growth rate over time. We observe that regret decreases as the re-solving frequency increases. This trend holds consistently across both algorithms and input types. In addition, while regret accumulates over longer time horizons, the increasing rate remains stable for algorithms employing higher re-solving frequencies. These findings are consistent with the guarantees of \cref{thm:spectrum}, as more frequent updates enable better adaptation to dynamic environments. 

\subsection{Comparative Analysis with Baseline Methods} \label{exp:2}

\begin{table*}[t]
    \centering
    \addtocounter{table}{-2}
    \caption{Regret of algorithms with various re-solving frequencies.
    \label{table:2}}
    \centering
    \resizebox{\textwidth}{!}{
    \begin{minipage}{0.57\textwidth}
    \centering
    \begin{tabular}{c | c c c c c}
        \toprule
        & $T$ & First-Order & Low freq & Mid freq & High freq \\
        \midrule
        \multirow{4}{*}{Input I} & $10^3$ & 12.13 & 7.76 & 5.77 & \textbf{4.86} \\
        & $10^4$ & 38.50 & 10.96 & 7.55 & \textbf{5.67} \\
        & $10^5$ & 122.44 & 23.90 & 9.92 & \textbf{8.36} \\
        & $10^6$ & 404.59 & 56.70 & 21.90 & \textbf{8.99} \\
        \midrule
        \multirow{4}{*}{Input II} & $10^3$ & 11.44 & 6.28 & 4.86 & \textbf{3.95} \\
        & $10^4$ & 36.50 & 10.21 & 7.34 & \textbf{3.81} \\
        & $10^5$ & 115.57 & 14.61 & 11.78 & \textbf{4.66} \\
        & $10^6$ & 365.99 & 35.20 & 15.68 & \textbf{6.26} \\
        \bottomrule
    \end{tabular}
    \vskip 0.08in
    \centering
    \subcaption{Algorithm \ref{alg:main-1}}
    \end{minipage} \quad
    \begin{minipage}{0.57\textwidth}
    \centering
    \begin{tabular}{c | c c c c c}
        \toprule
        & $T$ & First-Order & Low freq & Mid freq & High freq \\
        \midrule
        \multirow{4}{*}{Input I} & $10^3$ & 12.13 & 6.78 & 5.20 & \textbf{4.50} \\
        & $10^4$ & 38.50 & 10.37 & 8.03 & \textbf{5.99} \\
        & $10^5$ & 122.44 & 22.33 & 11.57 & \textbf{6.36} \\
        & $10^6$ & 404.59 & 48.21 & 22.44 & \textbf{7.09} \\
        \midrule
        \multirow{4}{*}{Input II} & $10^3$ & 11.44 & 3.20 & 2.56 & \textbf{1.75} \\
        & $10^4$ & 36.50 & 5.48 & 4.30 & \textbf{2.52} \\
        & $10^5$ & 115.57 & 12.35 & 4.48 & \textbf{3.86} \\
        & $10^6$ & 365.99 & 30.48 & 13.20 & \textbf{4.77} \\
        \bottomrule
    \end{tabular}
    \vskip 0.08in
    \centering
    \subcaption{Algorithm \ref{alg:main-2}}
    \end{minipage}
    }
\end{table*}

\begin{figure*}[h!]
    \centering
    \addtocounter{figure}{-2}
    \includegraphics[width=\linewidth]{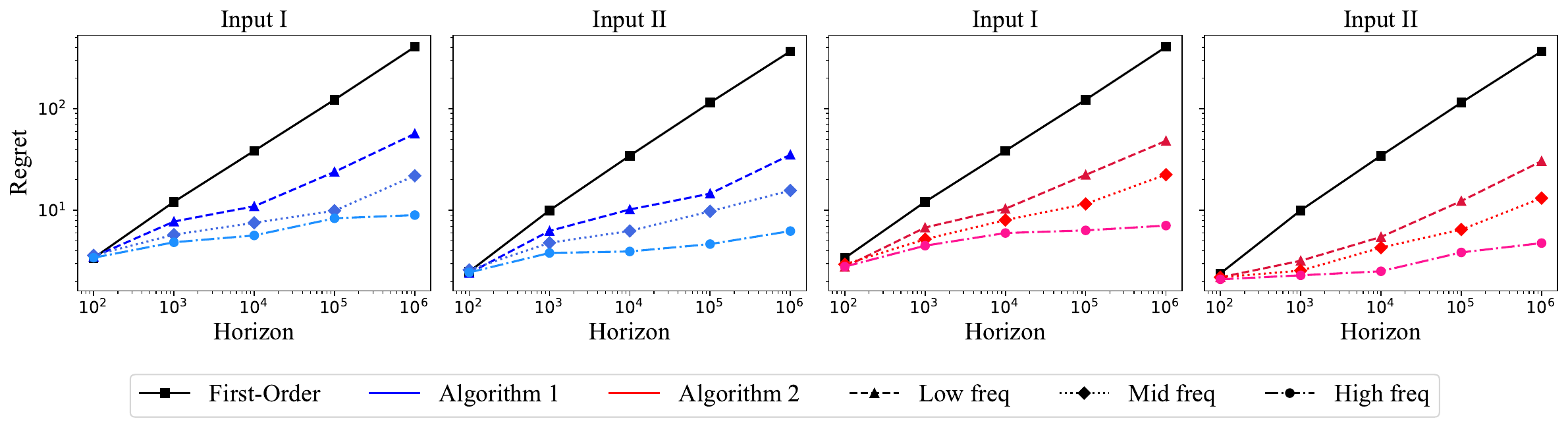}
    \vskip 0.08in
    \caption{Evaluations of \cref{alg:main-1} and \ref{alg:main-2} across various horizons, re-solving frequencies, and stochastic inputs, validating the positive relationship between regret and frequency stated in \cref{thm:spectrum}.
    \label{fig:1}}
    \vskip -0.06in
\end{figure*}

We next compare our algorithm's regret and computation time with a classic LP-based \cite{li2022online} and two first-order methods with $\mathcal{O}(T^{1/2})$ \cite{sun2020} and $\mathcal{O}(T^{1/3})$ regrets \cite{gao2024decoupling}. We generate $\{(r_t, \bm{a}_t)\}_{t=1}^T$ from a uniform distribution (Input I). We set the resource types to $m = 5$, time $T$ to range evenly over $[10^2, 10^6]$, average resource $d_i\sim \text{Uniform}[1/3, 2/3]$, and re-solving frequency to $f = T^{1/3}$ from Section \ref{exp:1}. Each result is averaged over 100 trial runs. We summarize the findings in Table \ref{table:4} and Figure \ref{fig:2}, and we include more comparisons with an infrequent re-solving method \cite{infrequent2024} in Appendix \ref{supp_exp:com}. We observe the effectiveness of \cref{alg:main-1} and \ref{alg:main-2}: 
\begin{enumerate}[leftmargin=10pt]
    \item Our algorithms exhibit strong performance in terms of decision optimality. They achieve over a 20-fold and 10-fold improvement in regret compared to the $\mathcal{O}(T^{1/2})$ first-order method and the $\mathcal{O}(T^{1/3})$ first-order method respectively. These numerical results also corroborate the theoretical bounds in Theorem \ref{thm:spectrum}. 
    \item Our algorithms are computationally efficient. Their runtimes exhibit over 100-fold improvement compared to the LP-based method with only a minimal increase (less than 2-fold) in regret.
\end{enumerate}
Therefore, we achieve a balance between effective decision-making and efficient computation. Our algorithms demonstrate better regret than the first-order method and obtain substantial computational speed-ups compared to the LP-based method. They exhibit strong scalability and adaptability across various re-solving frequencies and stochastic input models, consistently delivering superior performance as the problem size grows. 

\section{Conclusion and Discussion} 
\label{conclusion}

This paper presents a new approach to online linear programming for dynamic resource allocation, addressing the inherent trade-offs between decision-making optimality and computational efficiency. Recognizing the limitations of existing methods, we propose a parallel framework that decouples online learning and decision-making into independent yet complementary processes. By integrating LP-based and first-order methods, our framework effectively balances total regret and computational cost.

We establish rigorous theoretical guarantees, proving that our algorithm achieves a worst-case regret bound of $\mathcal{O}(\log(T/f) + \sqrt{f})$ under continuous support. This result highlights our method's ability to interpolate between LP-based and first-order methods based on computational capability. Furthermore, extensive experiments validate the effectiveness of our approach, demonstrating improvements in both regret minimization and runtime efficiency over competitive baselines.

Beyond these contributions, our work opens avenues for further research in adaptive online decision-making. Future directions include refining the re-solving frequency based on real-time computational constraints and extending our framework to broader classes of online optimization problems. Overall, our results underscore the potential of hybrid algorithms in high-dimensional, large-scale environments, offering practical insights for applications in operations research, machine learning, and beyond.


\subsection*{Discussion of Unknown Horizon}

In this paper, we consider decision-making under a finite horizon, since the average resource \cref{ass:1}(c) widely used in OLP literature relies on a fixed horizon to be well-defined. 
When the horizon is unknown, this assumption becomes ill-posed, and prior work \cite{balseiro2023online} shows that it may not be possible to achieve sublinear regret. 
Addressing these challenges would likely require first adapting LP-based and first-order OLP methods to uncertain horizons before tackling a hybrid approach. 

In practice, it is often possible to make a prior prediction of the horizon using data-driven approaches or based on the resources available. For example, in online advertising, the number of customers can be predicted from historical statistics. Besides, in retail or event-based sales, the selling horizon may be externally decided by upper-level decision-makers. These practical applications suggest a reasonable modeling choice for finite time horizons.

\newpage

\section*{Impact Statement}
This paper presents work whose goal is to advance the field of Machine Learning. There are many potential societal consequences of our work, none of which we feel must be specifically highlighted here.

\section*{Acknowledgments}
This work was supported in part by NSF grant CCF-2338226.

\renewcommand \thepart{}
\renewcommand \partname{}

\bibliography{olp.bib}
\bibliographystyle{icml2025}


\doparttoc
\faketableofcontents
\part{}

\newpage
\appendix
\onecolumn

\allowdisplaybreaks
\addcontentsline{toc}{section}{Appendix}
\part{Appendix} 
\parttoc

\paragraph{Structure of the Appendix} We organize the appendix as follows. In \textbf{Section \ref{app:foundation}}, we present some primary results for the OLP problem and the foundation analysis for LP-based and first-order methods; \textbf{Section \ref{app:framework}} characterizes the unique properties from our parallel multi-phase structure, and presents the cohesive mathematical framework which decomposes the performance metrics to unify all the OLP methods; \textbf{Section \ref{app:main}} demonstrates the outline of \cref{alg:main-2}, combines all the previous results and proves the main results in our paper; \textbf{Section \ref{app:addexp}} provides auxiliary results and technical support for the previous three sections; \textbf{Section \ref{app: supp_exp}} includes supplementary experiments of more general input distributions and comparisons with recent algorithms. 

\newpage
\section{Primary Results and Properties} \label{app:foundation}

In this section, we present some primary results for the problem and OLP dual algorithms which will help present our main results. We also include basic properties and convergence of the LP-based and first-order methods. All the analyses are stated under Assumption \ref{ass:1} and \ref{ass:2}. 

\subsection{Preliminary results}

We propose the following lemma as directed results from Assumptions. Let $\Xi$ denote the family of distributions satisfying \cref{ass:1} and \ref{ass:2}. We propose the following lemma.

\begin{lem}[Dimension Stability] \label{lem:delta}
    Under Assumptions \ref{ass:1} and \ref{ass:2}, there exists a constant $\delta > 0$ such that 
    \begin{align*}
        \forall \bm{d}_t \in \mathcal{D} \assign [d_i - \delta, d_i + \delta]^m, \text{ stochastic programs } \eqref{opt:d_dual} \text{ specified by } \bm{d}_t \text{ share the same } I_B \text{ and } I_N \text{ sets}.
    \end{align*}
\end{lem}

This lemma is a consequence of Lemmas 12 and 13 in \cite{li2022online}. The existence of $\delta$ comes from the continuity of $f_{\bm{d}_t}(\bm{p})$. Thus, $\delta$ is only associated with stochastic program \eqref{opt:d_dual}, and it is independent of $T$. Note that $\mathcal{D} \subset \mathcal{V}_{\bm{d}}$. We will analyze our feasible dual solutions derived from the online algorithm with $\bm{d}_t \in \mathcal{D}$. 

\begin{lem}[Quadratic Regularity, Proposition 2 in \cite{li2022online}]\label{lem:quad}
    Under Assumptions \ref{ass:1} and \ref{ass:2}, for any $\bm{p} \in \mathcal{V}_{\bm{p}}$, 
    \begin{align*}
        f(\bm{p}) & \leq f(\bm{p}^*) + \nabla f(\bm{p}^*)^\top (\bm{p} - \bm{p}^*) + \frac{\mu \bar{a}^2}{2} \lVert \bm{p} - \bm{p}^* \rVert^2, \\
        f(\bm{p}) & \geq f(\bm{p}^*) + \nabla f(\bm{p}^*)^\top (\bm{p} - \bm{p}^*) + \frac{\nu \lambda}{2} \lVert \bm{p} - \bm{p}^* \rVert^2.
    \end{align*}
    Moreover, $\bm{p}^*$ is the unique optimal solution to \eqref{opt:e_dual}.
\end{lem}

This lemma establishes a local form of semi-strong convexity and smoothness at $\bm{p}^*$, which is guaranteed by our assumptions on the distribution $\mathcal{P} \in \Xi$. By focusing on these local properties rather than insisting on strong convexity and global smoothness, we relax the classical requirements typically imposed in such settings. In later sections, we will leverage this result to derive our regret bound. 

\subsection{LP-based analysis} \label{app-lp}

We state the LP-based method in \cref{alg:lp-based} and its properties of boundedness and dual convergence results. 

\begin{algorithm}[h!]
\caption{LP-based method \label{alg:lp-based}}
\KwIn{total resource $\bm{b}$, time horizon $T$, average resource $\bm{d} = \bm{b} / T$, and initial dual price $\bm{p}_1 = \bm{0}.$}
\For{$t$ = \rm{$1$ to $T$ }}{

Observe $(r_t, \bm{a}_t)$ and make decision $x_t$ based on rule \eqref{decision_rule} if constraints are not violated. 

Update constraint for $i = 1, ..., m$: 

\hspace{15 pt} remaining resource constraint $b_{it} = b_{i, t-1} - a_{it} x_t$

\hspace{15 pt} average resource capacity $d_{it} = \frac{b_{it}}{T-t}$

Solve the updated dual problem and obtain dual price $\bm{p}_{t+1}$: \vspace{-5 pt}
\begin{align*}
    \bm{p}_{t+1} = \argmin_{\bm{p} \geq \bm{0}} \ \bm{d}_t^\top \bm{p} + \frac{1}{t} \sum_{j=1}^t (r_j - \bm{a}_j^\top \bm{p})^+
\end{align*}
\vspace{-5 pt}
}
\end{algorithm}

LP-based method incorporates the past decisions into the optimization of $\bm{p}_t$. If resources were over-utilized in earlier periods, the remaining supply decreases, prompting the algorithm to raise the dual price and become more selective with future orders. Conversely, if ample resources remain, the future dual price will be lowered, allowing more consumer requests to be accepted. This adaptive mechanism accounts for past actions by adjusting the available resource capacity.

\begin{lem}[Boundedness of LP result]
\label{lem:lp-bound}
    The online dual price $\bm{p}_t$ from Algorithm \ref{alg:lp-based} and the optimal dual price $\bm{p}^*$ of stochastic program \eqref{opt:e_dual} are bounded as
    \begin{align*}
        \lVert \bm{p}^* \rVert \leq \frac{\bar{r}}{\underline{d}}, \ \ \ \lVert \bm{p}_t \rVert \leq \frac{\bar{r}}{\underline{d} - \delta}.
    \end{align*}
\end{lem}

\begin{lem}[Dual Convergence of LP-based algorithm] \label{lem:conv-lp}
    Under Assumptions \ref{ass:1} and \ref{ass:2}, $\bm{p}_t$ represents the online solution from LP-based method, there exists a constant $C_{lp} > 0$ depending on $\bar{r}, \bar{a}, \underline{d}, m, \nu$, and $\lambda$ such that 
    \begin{align}
        \mathbb{E} \left[ \lVert \bm{p}_t - \bm{p}_{t}^* \rVert^2 \right] \leq \frac{C_{lp}}{t}.
    \end{align}

    In addition, the difference between $\bm{p}_t^*$ and $\bm{p}^*$ satisfies 
    \begin{align}
        \lVert \bm{p}_t^* - \bm{p}^* \rVert^2 \leq \frac{1}{\nu^2 \lambda^2} \lVert \bm{d}_t - \bm{d} \rVert^2.
    \end{align}
\end{lem}

This lemma establishes that the LP-based online AHDL algorithm \cite{jiang2024} produces dual solutions $\bm{p}_t$ that converge to $\bm{p}_t^*$. This convergence highlights the stability of the online dual variables during the intermediate stages of decision-making and ensures a warm start for the first-order method in the last re-solving batch. Moreover, we bound the distance between $\bm{p}_t^*$ and $\bm{p}^*$ by relating their respective average resource capacities, $\bm{d}_t$ and $\bm{d}$, in the associated stochastic programs \cite{li2022online}. These results provide the foundation for analyzing $\lVert \bm{p}_t - \bm{p}^* \rVert$ in our final result. 

\paragraph{Proof of \cref{lem:lp-bound}.} By the optimality of $\bm{p}^*$ and boundedness of $\bm{d}$, we have \begin{align*}
    \underline{d} \lVert \bm{p}^*\rVert_1 \leq \bm{d}^\top \bm{p}^* \leq \mathbb{E}[r] \leq \bar{r}.
\end{align*}
This holds because if otherwise, $\bm{p}^*$ can not be the optimal solution due to $f(\bm{p}^*) > f(\bm{0})$. 
Given the non-negativeness of $\bm{p}^*$ and $\bm{p}^* \in \mathcal{V}_{\bm{d}}$, we obtain $\lVert \bm{p}^* \rVert \leq \lVert \bm{p}^* \rVert_1$, and thus $\lVert \bm{p}^* \rVert \leq \frac{\bar{r}}{\underline{d}}$. \\

Similarly, by the optimality of $\bm{p}_t^*$ and its associated $\bm{d}_t \in \mathcal{D}$ in Lemma \ref{lem:delta}, we know \begin{align*}
    (\underline{d} - \delta) \lVert \bm{p}_t \rVert_1 \leq {\bm{d}}_t^\top \bm{p}_t \leq \mathbb{E}[r] \leq \bar{r},
\end{align*}
so we get the bound as $\lVert \bm{p}_t \rVert \leq \frac{\bar{r}}{\underline{d} - \delta}$. 

\paragraph{Proof of \cref{lem:conv-lp}.} According to the latest results in \cite{jiang2024}, the online dual price $\bm{p}_t$ achieves a sublinear convergence $\mathcal{O}(\frac{1}{\sqrt{t}})$ to $\bm{p}_t^*$. Since \begin{align*}
    & \bm{p}^* \in \arg\min_{\bm{p} \geq \bm{0}} f(\bm{p}) := \bm{d}^\top \bm{p} + \mathbb{E} \left[ (r - \bm{a}^\top \bm{p})^+ \right], \\
    & \bm{p}_t^* \in \arg\min_{\bm{p}_t \geq \bm{0}} f_{\bm{d}_t}(\bm{p}_t) := \bm{d}_t^\top \bm{p}_t + \mathbb{E} \left[ (r - \bm{a}^\top \bm{p}_t)^+ \right],
\end{align*}
By \textit{Lemma 12} in \cite{li2022online}, we have $\|\bm{p}_t^* - \bm{p}^*\|_2^2 \leq \frac{1}{\nu^2 \lambda^2} \|\bm{d}_t - \bm{d}\|_2^2$. 

\subsection{First-order Analysis} \label{app_fo}

We present the classic first-order method \cite{sun2020} in \cref{alg:first-order} and the first-order method with restart strategy \cite{gao2024decoupling} in \cref{alg:first-order_gao}. We also show their properties of boundedness and dual convergence results. 

\begin{algorithm}[h]
\caption{First-Order Online algorithm \label{alg:first-order}}	
\KwIn{total resource $\bm{b}$, time horizon $T$, average resource $\bm{d} = \bm{b} / T$, and initial dual price $\bm{p}_1 = \bm{0}.$}
\For{$t$ = \rm{$1$ to $T$ }}{

Observe $(r_t, \bm{a}_t)$ and make decision $x_t$ based on rule \eqref{eqn:kkt}. 

Update learning rate $\alpha_t = \mathcal{O}(1/\sqrt{T})$.

Compute subgradient and obtain dual price $\bm{p}_{t+1}$: 
\vspace{-5 pt}
\begin{align*}
    \bm{p}_{t+1} = \argmin_{\bm{p} \geq \bm{0}} \ (\bm{d} - \bm{a}_t x_t)^\top \bm{p} + \frac{1}{2\alpha_t} \lVert \bm{p} - \bm{p}_t \rVert^2
\end{align*}
\vspace{-5 pt}
}
\end{algorithm}

\begin{algorithm}[h]
\caption{First-Order Restart algorithm \label{alg:first-order_gao}}	
\KwIn{total resource $\bm{b}$, time horizon $T$, average resource $\bm{d} = \bm{b} / T$, and initial dual price $\bm{p}_1 = \bm{0}, \bm{p}^L_{1} = \bm{0}$}
\For{$t$ = \rm{$1$ to $T$ }}{

Observe $(r_t, \bm{a}_t)$ and make decision $x_t$ based on rule \eqref{decision_rule}

Update learning rate $\alpha_t = \begin{cases}
    \mathcal{O}(1/T^{1/3}) & t \leq T^{2/3} \\
    \mathcal{O}(1/T^{2/3}) & t > T^{2/3}
\end{cases}$

Compute subgradient and obtain dual price $\bm{p}_{t+1}$: 
\vspace{-5 pt}
\begin{align*}
    \bm{p}_{t+1} = \argmin_{\bm{p} \geq \bm{0}} \ (\bm{d} - \bm{a}_t x_t)^\top \bm{p} + \frac{1}{2\alpha_t} \lVert \bm{p} - \bm{p}_t \rVert^2
\end{align*}
\hspace{-7pt} Run subgradient method with stepsize $\alpha_t = \mathcal{O} (1/t)$ and update $\{\bm{p}_t^L\}$. At $t = T^{2/3}$, restart $\bm{p}_{T^{2/3}} = \bm{p}^L_{T^{2/3}}$.
}
\end{algorithm}

\begin{lem}[Boundedness of first-order result]
\label{lem:fo-bound}
    The online dual price $\bm{p}_t$ from Algorithm \ref{alg:first-order} is bounded as \begin{align*}
        \lVert \bm{p}_t \rVert \leq \frac{2\bar{r} + m(\bar{a}+\bar{d})^2}{\underline{d}} + m(\bar{a}+\bar{d}).
    \end{align*}
\end{lem}

\begin{lem}[Dual convergence of First-order algorithm]
\label{lem:conv-fo}
    Under Assumptions \ref{ass:1} and \ref{ass:2}, $\bm{p}_t$ represents the online solution from first-order method, if $\alpha_t < \nu \lambda$, the subgradient updates satisfy the following recursion rule:
    \begin{align}\label{eq3.3}
        \mathbb{E} \left[ \lVert \bm{p}_{t+1} - \bm{p}^* \rVert^2 \right] & \leq (1-\alpha_t \nu \lambda) \lVert \bm{p}_{t} - \bm{p}^* \rVert^2 + \alpha_t^2 m(\bar{a}+\bar{d})^2.
    \end{align}

    \hspace{3 pt} \textbf{Case 1.} if $\alpha_t \equiv \alpha < \frac{1}{\nu \lambda}$, then there exists a constant $C_{fo} = \frac{\bar{p}^2 + m(\bar{a}+\bar{d})^2}{\nu \lambda}$ such that 
    \begin{align}\label{eq3.3-case1}
        \mathbb{E} \left[ \lVert \bm{p}_t - \bm{p}^* \rVert^2 \right] \leq C_{fo} \left( \frac{1}{\alpha t} + \alpha \right).
    \end{align}
    
    \hspace{3 pt} \textbf{Case 2.} if $\alpha_t = \frac{2}{\nu \lambda (t+1)}$, then there exists a constant $C_{fo} = \frac{4m(\bar{a}+\bar{d})^2}{\nu^2 \lambda^2}$ such that 
    \begin{align}\label{eq3.3-case2}
        \mathbb{E} \left[ \lVert \bm{p}_t - \bm{p}^* \rVert^2 \right] \leq \frac{C_{fo}}{t}.
    \end{align}
\end{lem}

This lemma shows that the first-order method \cite{gao2024decoupling} guarantees the convergence of $\bm{p}_t$ to $\bm{p}^*$. It also highlights a key trade-off in choosing the learning rate $\alpha_t$, an aspect that will be central to our later optimality analysis. It also highlights a key trade-off in choosing the learning rate $\alpha_t$, an aspect that will be central to our later optimality analysis. With this groundwork in place, we now move on to bounding the total regret of our algorithm.

\paragraph{Proof of \cref{lem:fo-bound}.} As our initial choice $\bm{p}_1 = \bm{0}$, according to \textit{Lemma 1} in \cite{sun2020}, we get the above result. In addition, by \textit{Lemma B.1} in \cite{gao2024decoupling}, we have \begin{align*}
    \lVert \bm{p}_t \rVert \leq \frac{\bar{r}}{\underline{d}} + \frac{m(\bar{a}+\bar{d})^2 \alpha_t}{2 \underline{d}} + \alpha_t \sqrt{m}(\bar{a}+\bar{d}). 
\end{align*}

\paragraph{Proof of \cref{lem:conv-fo}.} Based on the updated rule in Algorithm \ref{alg:first-order}, we derive: \begin{align*}
    \lVert \bm{p}_{t+1} - \bm{p}^* \rVert^2 & \leq \lVert \bm{p}_t - \alpha_t (\bm{d} - \bm{a}_t x_t) - \bm{p}^* \rVert^2 \\
    &= \lVert \bm{p}_t - \bm{p}^* \rVert^2 - 2 \alpha_t \langle \bm{d} - \bm{a}_t x_t, \bm{p}_t - \bm{p}^* \rangle + \alpha_t^2 \lVert \bm{d} - \bm{a}_t x_t \rVert^2 \\
    & \leq \lVert \bm{p}_t - \bm{p}^* \rVert^2 - 2 \alpha_t \langle \bm{d} - \bm{a}_t x_t, \bm{p}_t - \bm{p}^* \rangle + \alpha_t^2 m (\bar{a} + \bar{d})^2.
\end{align*}

1) With convexity of $f$ and $\mathbb{E}[\bm{d} - \bm{a}_t x_t] \in \partial f(\bm{p}_t)$, we have: \begin{align*}
    f(\bm{p}^*) \geq f(\bm{p}_t) + \langle \bm{d} - \bm{a}_t x_t, \bm{p}_t - \bm{p}^* \rangle.
\end{align*}

2) With quadratic regularity of $f$ as in Lemma \ref{lem:quad}, we have: \begin{align*}
    & f(\bm{p}_t) \geq f(\bm{p}^*) + \nabla f(\bm{p}^*)^\top (\bm{p}_t - \bm{p}^*) + \frac{\nu \lambda}{2} \lVert \bm{p}_t - \bm{p}^* \rVert^2, \\
    \text{which indicates } \ \ & f(\bm{p}_t) - f(\bm{p}^*) \geq \frac{\nu \lambda}{2} \lVert \bm{p}_t - \bm{p}^* \rVert^2. 
\end{align*}

Combine the above results and take expectation conditioned on history information $\{(r_j, \bm{a}_j), j \leq t\}$, we obtain: \begin{align*}
    \mathbb{E}\lVert \bm{p}_{t+1} - \bm{p}^* \rVert^2 & \leq \lVert \bm{p}_t - \bm{p}^* \rVert^2 - 2 \alpha_t \mathbb{E}[\langle \bm{d} - \bm{a}_t x_t, \bm{p}_t - \bm{p}^* \rangle] + \alpha_t^2 m (\bar{a} + \bar{d})^2 \\
    & \leq \lVert \bm{p}_t - \bm{p}^* \rVert^2 - 2 \alpha_t (f(\bm{p}_t) - f(\bm{p}^*)) + \alpha_t^2 m (\bar{a} + \bar{d})^2 \\
    & \leq \lVert \bm{p}_t - \bm{p}^* \rVert^2 - \alpha_t \nu \lambda \lVert \bm{p}_t - \bm{p}^* \rVert^2 + \alpha_t^2 m (\bar{a} + \bar{d})^2 \\
    &= (1-\alpha_t \nu \lambda) \lVert \bm{p}_{t} - \bm{p}^* \rVert^2 + \alpha_t^2 m(\bar{a}+\bar{d})^2. 
\end{align*}
This proves \eqref{eq3.3} with the general case of learning rate $\alpha_t$. \\

\textbf{Case 1.} When $\alpha_t = \alpha < \frac{1}{\nu \lambda}$ is a constant, we take recursion of \eqref{eq3.3}. Note that $(1-\nu \lambda \alpha)^t < 1 / \nu \lambda \alpha t$, we get: \begin{align*}
    \mathbb{E}\lVert \bm{p}_{t+1} - \bm{p}^* \rVert^2 & \leq (1-\alpha \nu \lambda)^t \lVert \bm{p}_1 - \bm{p}^* \rVert^2 + \sum_{j=0}^{t-1} \alpha^2 m(\bar{a}+\bar{d})^2 (1-\alpha \nu \lambda)^j \\
    & \leq \frac{\lVert \bm{p}_1 - \bm{p}^* \rVert^2}{\nu \lambda \alpha t} + \frac{m(\bar{a}+\bar{d})^2}{\nu \lambda} \alpha.
\end{align*}
Since all feasible $\bm{p} \in \mathcal{V}_{\bm{p}}$ is bounded, let $C_{fo} = \frac{\bar{p}^2 + m(\bar{a}+\bar{d})^2}{\nu \lambda}$, we are able to obtain \eqref{eq3.3-case1}. \\

\textbf{Case 2.} When $\alpha_t = \frac{2}{\nu \lambda (t+1)}$, for any $j \leq t$, Lemma \ref{lem:conv-fo} gives us \begin{align*}
    \mathbb{E} \lVert \bm{p}_{j+1} - \bm{p}^* \rVert^2 \leq \frac{j-1}{j+1} \lVert \bm{p}_{j} - \bm{p}^* \rVert^2 + \frac{4m (\bar{a} + \bar{d})^2}{\nu^2 \lambda^2 (j+1)^2}.
\end{align*}

Re-arranging the above inequality, we have \begin{align*}
    & (j+1)^2 \mathbb{E} \lVert \bm{p}_{j+1} - \bm{p}^* \rVert^2 \leq (j^2 - 1) \lVert \bm{p}_{j} - \bm{p}^* \rVert^2 + \frac{4m (\bar{a} + \bar{d})^2}{\nu^2 \lambda^2}, \\
    \text{then } \ \ & (j+1)^2 \mathbb{E} \lVert \bm{p}_{j+1} - \bm{p}^* \rVert^2 - j^2 \lVert \bm{p}_{j} - \bm{p}^* \rVert^2 \leq \frac{4m (\bar{a} + \bar{d})^2}{\nu^2 \lambda^2}. 
\end{align*}

Then by telescoping from $j = 2$ to $t$, we have \begin{align*}
    \sum_{j=1}^t (j+1)^2 \mathbb{E} \lVert \bm{p}_{j+1} - \bm{p}^* \rVert^2 - j^2 \lVert \bm{p}_{j} - \bm{p}^* \rVert^2 = (t+1)^2 \mathbb{E} \lVert \bm{p}_{t+1} - \bm{p}^* \rVert^2 - \lVert \bm{p}_{1} - \bm{p}^* \rVert^2 \leq \frac{4m (\bar{a} + \bar{d})^2 t}{\nu^2 \lambda^2}
\end{align*}

which then gives us 
\begin{align*}
    & \mathbb{E} \lVert \bm{p}_{t+1} - \bm{p}^* \rVert^2 \leq \frac{\lVert \bm{p}_{1} - \bm{p}^* \rVert^2}{(t+1)^2} + \frac{4m (\bar{a} + \bar{d})^2 t}{\nu^2 \lambda^2 (t+1)^2}, \\
    \text{thus } \ \ & \mathbb{E} \lVert \bm{p}_{t} - \bm{p}^* \rVert^2 \leq \frac{\lVert \bm{p}_{1} - \bm{p}^* \rVert^2}{t^2} + \frac{4m (\bar{a} + \bar{d})^2}{\nu^2 \lambda^2 t}.
\end{align*}
As $\bm{p}_1 = \bm{0}$ and $\bm{p}^*$ is bounded, taking $C_{fo} = \frac{4m(\bar{a}+\bar{d})^2}{\nu^2 \lambda^2}$ completes the proof for \eqref{eq3.3-case2}.

\section{Algorithm Design and Analysis} \label{app:framework}

In this section, we present a cohesive mathematical framework to integrate the LP-based method and first-order method by developing a new performance metric. We also provide some unique properties for our algorithm design of parallel multi-phase structure. 

As stated in Algorithm \ref{alg:main-1}, we construct independent paths for online learning and online decision-making conducted simultaneously. Specifically, for the online learning path, we use the LP-based method to resolve updated linear programs at a fixed frequency and send this result to the decision-making path; For the online decision-making path, we apply the first-order method in the initial and final batches, and use the latest dual price for decision-making in the intermediate resolving intervals. This process is illustrated in \cref{fig:alg1}.

\subsection{Regret Analysis}

First, we construct an upper bound for the offline optimal objective value. The challenge comes from the intractable dependency of constraints on objective value under the online setting. We tackle this issue by introducing a Lagrangian function to integrate constraints into the objective and balance revenue maximization with constraint satisfaction. The formulation is stated as follows. 

\begin{lem}[Lagrangian Upper Bound]
\label{lem:lagrangian}
    Under Assumptions \eqref{ass:1} and \eqref{ass:2}, define the deterministic Lagrangian dual function as \begin{align*}
        \ell (\bm{p}) \assign \mathbb{E} \left[rI(r > \bm{a}^\top \bm{p}) + (\bm{d} - \bm{a} I(r > \bm{a}^\top \bm{p}))^\top \bm{p}^* \right].
    \end{align*}
    Then for any feasible $\bm{p} \in \mathcal{V}_{\bm{p}}$, we have: \begin{align}
        (a) \quad & \mathbb{E} \bigg[\sum_{t=1}^T r_t x_t^* \bigg] \leq T \ell(\bm{p}^*), \nonumber \\
        (b) \quad & \ell(\bm{p}^*) - \ell(\bm{p}) \leq \mu \bar{a}^2 \lVert \bm{p} - \bm{p}^* \rVert^2.
    \end{align}
\end{lem}

\begin{lem}[Dual Price Boundedness] 
\label{lem:dual-bound}
    Under Assumption \ref{ass:1}, the online and offline optimal dual prices are bounded respectively by $\lVert\bm{p}^* \rVert \leq \tfrac{\bar{r}}{\underline{d}}$ and  
    \[\lVert \bm{p}_t \rVert \leq \bar{p} \assign \max \bigg(\frac{\bar{r}}{\underline{d} - \delta}, \ \frac{2\bar{r} + m(\bar{a}+\bar{d})^2}{\underline{d}} + m(\bar{a}+\bar{d}) \bigg).\]
\end{lem}

This lemma establishes that the optimal dual prices remain bounded. Our algorithm maintains these bounds because if $\bm{p}_t$ grows large, the algorithm responds by accepting more orders, which in turn reduces $\bm{p}_{t+1}$. This self-correcting mechanism keeps the dual prices within these limits. 
With this groundwork in place, we now move on to derive the upper bound for the total performance of \cref{alg:main-1}. 

\begin{thm}[Decomposition of Regret]
\label{thm:gap}
    Under Assumptions \eqref{ass:1} and \eqref{ass:2}, let $C_r = \max\{\frac{\bar{r}}{\underline{d}}, \frac{\nu \lambda \bar{a}^2}{2} \}$, we derive an upper bound for the regret $r(\bm{x})$ as: \begin{align}
        r(\bm{x}) \leq C_r \bigg[ \mathbb{E} \Big\lVert \Big(\bm{b} - \sum_{t=1}^T \bm{a}_t x_t \Big)^{B^+} \Big\rVert + \sum_{t=1}^T \mathbb{E} \lVert \bm{p}_t - \bm{p}^* \rVert^2 \bigg]. 
    \end{align}
\end{thm}

\paragraph{Proof of \cref{lem:lagrangian}.} This lemma is proved as \textit{Lemma 3} in \cite{li2022online} by strong duality and optimality of $\bm{p}_T^*$. 

\paragraph{Proof of Lemma \ref{lem:dual-bound}.} 
By the boundedness results in Lemma \ref{lem:lp-bound} and Lemma \ref{lem:fo-bound}, define \begin{align*}
    \bar{p} \assign \max \left(\frac{\bar{r}}{\underline{d} - \delta}, \ \frac{2\bar{r} + m(\bar{a}+\bar{d})^2}{\underline{d}} + m(\bar{a}+\bar{d}) \right),
\end{align*}
then we complete the proof. 

\paragraph{Proof of \cref{thm:gap}.} By the definition of $\ell(\bm{p})$, we derive the online objective as \begin{align*}
    \mathbb{E} \bigg[\sum_{t=1}^T r_t x_t \bigg] &= \sum_{t=1}^T \mathbb{E}[\mathbb{E}[r_t x_t | \bm{p}_t]] \\
    &= \sum_{t=1}^T \mathbb{E}[\ell(\bm{p}_t) - (\bm{d} - \bm{a}_t x_t)^\top \bm{p}^*].
\end{align*}

Then by Lemma \ref{lem:lagrangian}, we derive the upper bound for the regret $r(\bm{x})$ as defined in \eqref{def:regret} to be: \begin{align*}
    r(\bm{x}) &= \mathbb{E} \bigg[\sum_{t=1}^T r_t x_t^* - r_t x_t \bigg] \\
    & \leq T \ell(\bm{p}^*) - \sum_{t=1}^T \mathbb{E}[\ell(\bm{p}_t) - (\bm{d} - \bm{a}_t x_t)^\top \bm{p}^*] \\
    &= \sum_{t=1}^T \mathbb{E}[(\bm{d} - \bm{a}_t x_t)^\top \bm{p}^*] + \sum_{t=1}^T \mathbb{E}[\ell(\bm{p}^*) - \ell(\bm{p}_t)] \\
    & \leq \mathbb{E} \bigg[ \Big(\bm{b} - \sum_{t=1}^T \bm{a}_t x_t \Big)^\top \bm{p}^* \bigg] + \frac{\mu \bar{a}^2}{2} \sum_{t=1}^T \mathbb{E} \lVert \bm{p}_t - \bm{p}^* \rVert^2 \\
    & \leq \lVert \bm{p}^* \rVert \cdot \mathbb{E} \bigg\lVert \Big(\bm{b} - \sum_{t=1}^T \bm{a}_t x_t \Big)^{B^+} \bigg\rVert + \frac{\mu \bar{a}^2}{2} \sum_{t=1}^T \mathbb{E} \lVert \bm{p}_t - \bm{p}^* \rVert^2
\end{align*}
where $( \cdot )^{B+}$ denotes the positive part only for binding constraints. 

As $\lVert \bm{p}^* \rVert \leq \frac{\bar{r}}{\underline{d}}$ is bounded, taking the constant $C_r = \max\{\frac{\bar{r}}{\underline{d}}, \frac{\mu \bar{a}^2}{2} \}$ completes the proof.

\subsection{Total Performance Analysis}

We consider the constraint violation as $v(\bm{x}) = \lVert (\bm{Ax} - \bm{b})^+ \rVert$ in \eqref{def:vio}. Then the total performance is the combination of regret and constraint violation. Since Theorem \ref{thm:gap} holds for any unknown distribution $\mathcal{P} \in \Xi$, we have: 
\begin{align} \label{eq:total-regret}
    \Delta_T &= \sup_{\mathcal{P} \in \Xi} \mathbb{E}_{\mathcal{P}}[r(\bm{x}) + v(\bm{x})] \nonumber \\
    & \leq C_r \bigg[ \mathbb{E} \Big\lVert \Big(\bm{b} - \sum_{t=1}^T \bm{a}_t x_t \Big)^{B^+} \Big\rVert + \sum_{t=1}^T \mathbb{E} \lVert \bm{p}_t - \bm{p}^* \rVert^2 \bigg] + \mathbb{E} \Big\lVert \Big(\sum_{t=1}^T \bm{a}_t x_t - \bm{b} \Big)^+ \Big\rVert \nonumber \\
    & \leq C_r \bigg[\sum_{t=1}^T \mathbb{E} \lVert \bm{p}_t - \bm{p}^* \rVert^2 + \mathbb{E} \big\lVert (\bm{b} - \bm{Ax})^{B^+} \big\rVert + \mathbb{E} \big\lVert (\bm{Ax} - \bm{b})^+ \big\rVert \bigg]. 
\end{align}

This new framework is adaptive to all OLP methods. We proceed with this structure to analyze various OLP methods. 

\begin{thm}[Horizon Division] \label{thm:division}
    Based on the change of methods, we separate the horizon into three intervals $T_1, T_2$, and $T_3$, and reorganize the total performance for analysis to be \begin{align}
        \Delta_T \leq C_r \big[ \Delta_{T_1} + \Delta_{T_2} + \Delta_{T_3} \big].
    \end{align}
    where $\Delta_{T_1}, \Delta_{T_2}$, and $\Delta_{T_3}$ are the performances for each interval. 
\end{thm}

\paragraph{Proof of \cref{thm:division}.} We split the total horizon into three intervals of initial batch $T_1 = [0, f]$, intermediate process $T_2 = [f, kf]$, and final batch $T_3 = [kf, T]$. We use the first-order method for the first and final batches and the LP-based method for intermediate processes. Notice that for the initial resolving batch, we maintain the classical analysis for the first-order method since it has not reached the mixture with the LP-based method. Then based on \eqref{eq:total-regret}, we define 
\begin{align} \label{eq:split}
    \Delta_{T_1} & \assign \sum_{t=1}^f \mathbb{E} [r_t x_t^* - r_t x_t] + \mathbb{E} \Big\lVert \Big(\sum_{t=1}^f \bm{a}_t x_t - f\bm{d} \Big)^+ \Big\rVert \nonumber \\
    \Delta_{T_2} & \assign \sum_{t=f}^{kf} \mathbb{E} \lVert \bm{p}_t - \bm{p}^* \rVert^2 + \mathbb{E} \Big\lVert \Big((k-1)f \bm{d} - \sum_{t=f}^{kf} \bm{a}_t x_t \Big)^{B^+} \Big\rVert + \mathbb{E} \Big\lVert \Big(\sum_{t=f}^{kf} \bm{a}_t x_t - (k-1)f \bm{d} \Big)^+ \Big\rVert \\
    \Delta_{T_3} & \assign \sum_{t=kf}^T \mathbb{E} \lVert \bm{p}_t - \bm{p}^* \rVert^2 + \mathbb{E} \Big\lVert \Big((T-kf) \bm{d} - \sum_{t=kf}^T \bm{a}_t x_t \Big)^{B^+} \Big\rVert + \mathbb{E} \Big\lVert \Big(\sum_{t=kf}^T \bm{a}_t x_t - (T-kf) \bm{d} \Big)^+ \Big\rVert \nonumber
\end{align}

Then we have \begin{align*}
    \Delta_T \leq C_r \big[ \Delta_{T_1} + \Delta_{T_2} + \Delta_{T_3} \big]
\end{align*}
since $\lVert (a + b)^+ \rVert \leq \lVert a^+ \rVert + \lVert b^+ \rVert$ for any $a, b$.

\subsection{Regret for LP-based Method}

In this section, we are going to bound $\Delta_{T_2}$ for the regret from LP-based method. To provide a tighter analysis, we examine the real-time average resource capacity $\bm{d}_t$ instead of the original process $\bm{b}_t$. Specifically, by Lemma \ref{lem:delta}, we constraint $\bm{d}_t$ within a feasible range $\mathcal{D}$. If $\bm{d}_t \notin \mathcal{D}$, we set the dual price to zero to accept all subsequent orders. We analyze the remaining resources and constraint violation due to this strategy. Note that this is required solely for rigor analysis purposes and is not necessary when running the algorithm. 

\begin{lem}[Dynamics of Resource Usage] \label{lem:d-change}
    Under Assumption \ref{ass:1} and \ref{ass:2}, there exists a constant $C > 0$ depending on $\bar{d}, \bar{a}, m, \nu, \lambda_1, \mu$, and $C_{lp}$ such that \begin{align}
        \sum_{t=f}^{kf} \mathbb{E}[(d_{i, t} - d_i)^2] \leq C \log(k).
    \end{align}
\end{lem}

\begin{thm}[Regret of Intermediate Process]
\label{thm:lp-regret}
    Under Assumption \ref{ass:1} and \ref{ass:2}, following the result in \eqref{eq:split}, we prove the regret satisfies \begin{align}
        \Delta_{T_2} \leq \log \Big( \frac{T}{f} \Big) = \mathcal{O}(\log(k)).
    \end{align}
\end{thm}

\paragraph{Proof of \cref{lem:d-change}.} During the re-solving process, we denote each interval to be $[jf, (j+1)f]$ where $j = 1, 2, ..., k$. As the resource usage follows $\bm{b}_{t+1} = \bm{b}_t - \bm{a}_{t+1} I(r_{t+1} > \bm{a}_{t+1}^T \bm{p}_{{t+1}})$, normalizing both sides, we derive the update of average resource consumption to be: \begin{align*}
    d_{i, {(j+1)f}} = d_{i, jf} + \frac{\sum_{\ell = jf+1}^{(j+1)f} d_{i, jf} - {a}_{i, \ell} I(r_{\ell} > \bm{a}_{\ell}^\top \bm{p}_{(k+1)f})}{T - (j+1)f}.
\end{align*}

Subtracting $d_i$ on both sides, it becomes \begin{align*}
    d_{i, (j+1)f} - d_i &= {d}_{i, jf} - {d}_i + \frac{\sum_{\ell = jf+1}^{(j+1)f} {d}_{i, jf} - {a}_{i, \ell} I(r_{\ell} > \bm{a}_{\ell}^\top \bm{p}_{(j+1)f})}{T - (j+1)f} \\
    &= {d}_{i, jf} - {d}_i + \frac{\sum_{\ell = jf+1}^{(j+1)f} {d}_{i, jf} - {a}_{i, \ell} I(r_{\ell} > \bm{a}_{\ell}^\top \bm{p}_{jf}^*)}{T - (j+1)f} \\
    & \ \ + \frac{\sum_{\ell = jf+1}^{(j+1)f} {a}_{i, \ell} I(r_{\ell} > \bm{a}_{\ell}^\top \bm{p}_{jf}^*) - {a}_{i, \ell} I(r_{\ell} > \bm{a}_{\ell}^\top \bm{p}_{(j+1)f})}{T - (j+1)f}.
\end{align*}

Taking expectations of squares, we have \begin{align} \label{eq:expand}
     \mathbb{E} (d_{i, (j+1)f} - d_i)^2 = & \mathbb{E} (d_{i, jf} - d_i)^2 + \mathbb{E} \left[\frac{(\sum_{\ell = jf+1}^{(j+1)f} d_{i, jf} - {a}_{i, \ell} I(r_{\ell} > \bm{a}_{\ell}^\top \bm{p}_{jf}^*))^2}{(T - (j+1)f)^2} \right] \nonumber \\
    + & \mathbb{E} \left[\frac{(\sum_{\ell = jf+1}^{(j+1)f} {a}_{i, \ell} I(r_{\ell} > \bm{a}_{\ell}^\top \bm{p}_{jf}^*) - {a}_{i, \ell} I(r_{\ell} > \bm{a}_{\ell}^\top \bm{p}_{(j+1)f}))^2}{(T - (j+1)f)^2} \right] \nonumber \\
    + 2 & \mathbb{E} \left[(d_{i, jf} - d_i) \left(\frac{\sum_{\ell = jf+1}^{(j+1)f} d_{i, jf} - {a}_{i, \ell} I(r_{\ell} > \bm{a}_{\ell}^\top \bm{p}_{jf}^*)}{T - (j+1)f} \right) \right] \nonumber \\
    + 2 & \mathbb{E} \left[(d_{i, jf} - d_i) \left(\frac{\sum_{\ell = jf+1}^{(j+1)f} {a}_{i, \ell} I(r_{\ell} > \bm{a}_{\ell}^\top \bm{p}_{jf}^*) - {a}_{i, \ell} I(r_{\ell} > \bm{a}_{\ell}^\top \bm{p}_{(j+1)f})}{T - (j+1)f} \right) \right] \nonumber \\
    + 2 & \mathbb{E} \Bigg[\frac{\sum_{\ell = jf+1}^{(j+1)f} d_{i, jf} - {a}_{i, \ell} I(r_{\ell} > \bm{a}_{\ell}^\top \bm{p}_{jf}^*)}{T - (j+1)f} \cdot \frac{\sum_{\ell = jf+1}^{(j+1)f} {a}_{i, \ell} I(r_{\ell} > \bm{a}_{\ell}^\top \bm{p}_{jf}^*) - {a}_{i, \ell} I(r_{\ell} > \bm{a}_{\ell}^\top \bm{p}_{(j+1)f})}{T - (j+1)f} \Bigg].
\end{align}

By Lemma \ref{aux:dynamic}, we obtain the recursion relation as \begin{align} \label{eq:rec}
     \mathbb{E} (d_{i, (j+1)f} - d_i)^2 & \leq \mathbb{E} ({d}_{i, jf} - d_i)^2 + \frac{C_{rec}}{(k - j - 1)^2 f} + \frac{4\mu \bar{a}^2 \sqrt{C_{lp}}}{(k - j -1) \sqrt{(j+1)f}} \sqrt{\mathbb{E}[({d}_{i, jf} - d_i)^2]}
\end{align}
where $C_{rec} > 0$ is a constant defined in Lemma \ref{aux:dynamic}. 

Then according to Lemma \ref{aux:sum-d}, take $C = 12 \max\{ C_{rec}, 16\mu^2 \bar{a}^4 C_{lp}\}$, we solve recursion \eqref{eq:rec} and obtain the upper bound of total deviation from original $d_i$ in the re-solving process to be: \begin{align}\label{eq:outer}
    \sum_{j=1}^{k} \mathbb{E} \left[ ({d}_{i, jf} - {d}_i)^2 \right] & \leq \frac{C}{f} \log(k).
\end{align}

Therefore, we sum the whole re-solving process and obtain: \begin{align}
    \sum_{t=f}^{kf} \mathbb{E}[(d_{i, t} - d_i)^2] &= \sum_{j=1}^k \sum_{\ell=jf}^{(j+1)f} \mathbb{E}[(d_{i, \ell} - d_i)^2] \nonumber \\
    &= f \cdot \frac{C}{f} \log(k) \ \ \text{ (by result in \eqref{eq:outer}} \nonumber \\
    & \leq C \log(k).
\end{align}
This completes the proof. 

\paragraph{Proof of \cref{thm:lp-regret}.} We analyze the three terms in \eqref{eq:split} respectively. \begin{enumerate}
    \item For dual convergence, by Lemma \ref{lem:conv-lp}, we have \begin{align}\label{eq:lp-dual}
        \sum_{t=f}^{kf} \mathbb{E} \lVert \bm{p}_t - \bm{p}^* \rVert^2 &= \mathbb{E} \left[\sum_{j=1}^{k-1} \sum_{t=jf+1}^{(j+1)f} \lVert \bm{p}_t - \bm{p}^* \rVert_2^2 \right] \nonumber \\
        &= \sum_{j=1}^{k-1} f \ \mathbb{E} \Big[ \lVert \bm{p}_{(j+1)f} - \bm{p}^* \rVert_2^2 \Big] \nonumber \\
        & \leq \sum_{j=1}^{k-1} f \ \mathbb{E} \Big[ \lVert \bm{p}_{(j+1)f} - \bm{p}_{jf}^* \rVert_2^2 + \lVert \bm{p}_{jf}^* - \bm{p}^* \rVert_2^2 \Big] \nonumber \\
        & \leq \sum_{j=1}^{k-1} f \Big[\frac{C_{lp}}{jf} + \frac{1}{\nu^2 \lambda^2} \ \mathbb{E} \big[ (\bm{d}_{jf} - \bm{d})^2 \big] \Big] \ \ \ \text{(by Lemma \ref{lem:conv-lp})} \nonumber \\
        & \leq C_{lp} \log(k) + \frac{m C}{\nu^2 \lambda^2} \log(k). \ \ \ \text{(by Lemma \ref{lem:d-change})}. 
    \end{align}
    \item For the constraint violation, by Chebyshev's inequality, we bound the probability as \begin{align} \label{eq:prob}
        \sum_{t=f}^{kf} \mathbb{P}(|d_{i, t} - d_i| \leq \delta) & \leq \sum_{t=f}^{kf} \frac{\mathbb{E}[(d_{i, t} - d_i)^2]}{\delta^2} \nonumber \\
        & \leq \frac{C}{\delta^2} \log(k). \ \ \text{ (by Lemma \ref{lem:d-change})}
    \end{align}
    Since our strategy is to automatically set $\bm{p}_t = 0$ once $\bm{d}_t \notin \mathcal{D}$, we will accept all the subsequent orders according to our decision condition \eqref{eqn:kkt}. Denote $R$ as the total number of orders processed specifically due to this rule, then we have:
    \begin{align}\label{eq:stopping-time}
        \mathbb{E}[R] \leq \sum_{t=f}^{kf} \mathbb{P}(|d_{i, t} - d_i| \leq \delta) \leq \frac{C}{\delta^2} \log(k)
    \end{align}
    which indicates that the resource violation is at most \begin{align}\label{lp-violation}
        \mathbb{E} \Big\lVert \Big(\sum_{t=f}^{kf} \bm{a}_t x_t - (k-1)f \bm{d} \Big)^+ \Big\rVert \leq \mathbb{E}[\bar{a} R] \leq \frac{C \bar{a}}{\delta^2} \log(k).
    \end{align}
    \item For the remaining resource, we derive its upper bound by considering the opposite extreme case of our strategy. If we reject all those $R$ orders, then no resource is used on them. Thus, by \eqref{eq:stopping-time}, the positive projection of remaining resource at $t = kf$ is at most:
    \begin{align} \label{eq:lp-binding}
        \mathbb{E} \Big\lVert \Big((k-1)f \bm{d} - \sum_{t=f}^{kf} \bm{a}_t x_t \Big)^{B^+} \Big\rVert \leq \mathbb{E}[(d_i + \delta) R] \leq \frac{C (\bar{d} + \delta)}{\delta^2} \log(k). 
    \end{align}
\end{enumerate}

Combining the results of \eqref{eq:lp-dual}, \eqref{lp-violation}, and \eqref{eq:lp-binding}, we derive the final result as \begin{align}\label{eq:lp-total-regret}
    \Delta_{T_2} \leq \Big(C_{lp} + \frac{m C}{\nu^2 \lambda^2} + (\bar{a} + \bar{d} + \delta) \frac{C}{\delta^2} \Big) \log(k).
\end{align}
This completes the proof. 



\subsection{Regret for First-Order Method}

In this section, we are going to bound $\Delta_{T_1}$ and $\Delta_{T_3}$ for the regret from the first-order method. The key point is to analyze the connection between dual prices based on the gradient-updated rule. Here $\bm{p}_t = \bm{p}_t^{fo}$.

\begin{thm}[Regret of Initial Batch] 
\label{thm:fo-regret-1}
    Under Assumption \ref{ass:1} and \ref{ass:2}, we have \begin{align}
        \Delta_{T_1}\leq \left(\frac{m(\bar{a} + \bar{d})^2}{2} + \frac{2\bar{r} + m(\bar{a}+\bar{d})^2}{\underline{d}} + m(\bar{a}+\bar{d}) \right) \sqrt{f}.
    \end{align}
\end{thm}

\begin{lem}[Constraint Violation] \label{lem:vio}
    Under Assumption \ref{ass:1} and \ref{ass:2}, take the learning rate $\alpha_t = \alpha$, we bound the constraint violation for the last batch as \begin{align}
        \mathbb{E} \Big\lVert \Big(\sum_{t=kf}^T \bm{a}_t x_t - (T-kf) \bm{d} \Big)^+ \Big\rVert \leq \Big( \frac{C_{lp}}{\sqrt{kf}} + \frac{C_b}{T^2} \Big) \cdot \frac{1}{\alpha} + \frac{C_b}{\sqrt{(T-kf) kf}} \cdot \frac{1}{\alpha \sqrt{\alpha}} + \frac{C_b}{\sqrt{\alpha}}.
    \end{align}
\end{lem}

\begin{thm}[Regret of Final Batch] 
\label{thm:fo-regret-2}
    Under Assumption \ref{ass:1} and \ref{ass:2}, we have \begin{align}
        \Delta_{T_3} \leq \bigg(\frac{m(\bar{a}+\bar{d})^2}{\nu \lambda} + \frac{2C_{lp}}{\sqrt{k}} + \frac{C_{lp} \log(f)}{\nu \lambda k} + \frac{4C_b}{\sqrt{k}} + 2C_b \bigg) \cdot f^{1/3}. 
    \end{align}
\end{thm}

\begin{lem}[Warm-Start First-Order Regret]
\label{lem:fo-special}
    Assume that we have a warm start for the initial dual price $\bm{p}_0$ in the first batch, which satisfies $\lVert \bm{p}_0 - \bm{p}^* \rVert \leq f^{-1/3}$. Then the regret of the first batch follows \begin{align}
        \Delta_{T_1 \text{warm}} \leq \Big( \frac{m(\bar{a}+\bar{d})^2}{\nu \lambda} + 4C_w \Big) \cdot f^{1/3} + \log(f). 
    \end{align}
\end{lem}

\paragraph{Proof of \cref{thm:fo-regret-1}.} We analyze $\Delta_{T_1}$ in \eqref{eq:split} by parts. By \textit{Lemma B.3, B.4} in \cite{gao2024decoupling}, we have \begin{align*}
    & \sum_{t=1}^f \mathbb{E} [r_t x_t^* - r_t x_t] \leq \frac{m(\bar{a} + \bar{d})^2}{2} \alpha f, \nonumber \\
    & \mathbb{E} \Big\lVert \Big(\sum_{t=1}^f \bm{a}_t x_t - f\bm{d} \Big)^+ \Big\rVert \leq \frac{1}{\alpha} \bigg[\frac{2\bar{r} + m(\bar{a}+\bar{d})^2}{\underline{d}} + m(\bar{a}+\bar{d})\bigg].
\end{align*}

To achieve the tight upper bound, we select the optimal step size $\alpha = \frac{1}{\sqrt{f}}$, which then gives us the desired result. 

\paragraph{Proof of \cref{lem:vio}.} With the update rule of first-order method in Algorithm \ref{alg:first-order}, we have \begin{align*}
    & \bm{p}_{t+1} = [\bm{p}_t - \alpha (\bm{d} - \bm{a}_t x_t)]^+ \geq \bm{p}_t - \alpha (\bm{d} - \bm{a}_t x_t), \\
    \text{which gives us } \ \ & \bm{a}_t x_t - \bm{d} \leq \frac{1}{\alpha} (\bm{p}_{t+1} - \bm{p}_t).
\end{align*}
Summarizing on both sides and applying the telescoping, we derive \begin{align*}
    \sum_{t=kf}^T (\bm{a}_t x_t - \bm{d}) \leq \frac{1}{\alpha} \sum_{t=kf}^T (\bm{p}_{t+1} - \bm{p}_t) = \frac{1}{\alpha} (\bm{p}_{T+1} - \bm{p}_{kf}).
\end{align*}

By Lemma \ref{lem:fo-rec}, take $C_b = \max \{\frac{C_{lp}}{\sqrt{\nu \lambda}}, \frac{m(\bar{a} + \bar{d})}{\sqrt{\nu \lambda}}, \bar{p} \}$, we have 
\begin{align}\label{eq:prob-bound}
    \mathbb{E} \lVert \bm{p}_{T+1} - \bm{p}^* \rVert \leq C_b \Big[ \frac{1}{\sqrt{(T-kf) kf}} \cdot \frac{1}{\sqrt{\alpha}} + \sqrt{\alpha} + \frac{1}{T^2} \Big]. 
\end{align}

Thus, according to Lemma \ref{lem:conv-lp} and \eqref{eq:prob-bound}, the expectation of constraint violation satisfies 
\begin{align}\label{eq:vio-dev}
    \mathbb{E} \Big\lVert \Big(\sum_{t=kf}^T (\bm{a}_t x_t - \bm{d}) \Big)^+ \Big\rVert & \leq \mathbb{E} \Big\lVert \sum_{t=kf}^T (\bm{a}_t x_t - \bm{d}) \Big\rVert \nonumber \\
    & \leq \frac{1}{\alpha} \mathbb{E} \lVert \bm{p}_{T+1} - \bm{p}_{kf} \rVert \nonumber \\
    & \leq \frac{1}{\alpha} \mathbb{E} \Big[\lVert \bm{p}_{kf} - \bm{p}^* \rVert + \lVert \bm{p}_{T+1} - \bm{p}^* \rVert \Big] \nonumber \\
    & \leq \frac{1}{\alpha} \Big(\frac{C_{lp}}{\sqrt{kf}} + C_b \Big[ \frac{1}{\sqrt{(T-kf) kf}} \cdot \frac{1}{\sqrt{\alpha}} + \sqrt{\alpha} + \frac{1}{T^2} \Big] \Big) \nonumber \\
    & \leq \frac{C_{lp}}{\sqrt{kf}} \cdot \frac{1}{\alpha} + \frac{C_b}{\sqrt{(T-kf) kf}} \cdot \frac{1}{\alpha \sqrt{\alpha}} + C_b \cdot \frac{1}{\sqrt{\alpha}} + \frac{C_b}{T^2} \cdot \frac{1}{\alpha}. 
\end{align}
This completes the proof. 

\paragraph{Proof of \cref{thm:fo-regret-2}.} We decompose $\Delta_{T_3}$ into three parts according to \eqref{eq:split} and analyze each term respectively. Since the first-order method re-starts from $\bm{p}_{kf}$, by Lemma \ref{lem:conv-lp}, we have: \begin{align} \label{eq:fo-dual}
    \sum_{t=kf}^T \mathbb{E} \lVert \bm{p}_t - \bm{p}^* \rVert^2 & \leq \sum_{t=kf}^T \Big[ \frac{\lVert \bm{p}_{kf} - \bm{p}^* \rVert^2}{\nu \lambda \alpha t} + \frac{m(\bar{a}+\bar{d})^2}{\nu \lambda} \alpha \Big] \nonumber \\
    & \leq \frac{\lVert \bm{p}_{kf} - \bm{p}^* \rVert^2}{\nu \lambda \alpha} \log(T-kf) + \frac{m(\bar{a}+\bar{d})^2}{\nu \lambda} (T-kf) \alpha \nonumber \\
    & \leq \frac{\lVert \bm{p}_{kf} - \bm{p}^* \rVert^2}{\nu \lambda \alpha} \log(f) + \frac{m(\bar{a}+\bar{d})^2}{\nu \lambda} f \alpha \nonumber \\
    & \leq \frac{C_{lp} \log(f) }{\nu \lambda kf} \cdot \frac{1}{\alpha} + \frac{m(\bar{a}+\bar{d})^2}{\nu \lambda} f \alpha.
\end{align}

Since $t \in [kf, T]$, by \textit{Proposition 3.3} in \cite{gao2024decoupling}, the dual price is far from the origin and thus the positive projection still equals itself. This indicates that: \begin{align*}
    & \bm{p}_{t+1} = [\bm{p}_t - \alpha (\bm{d} - \bm{a}_t x_t)]^+ = \bm{p}_t - \alpha (\bm{d} - \bm{a}_t x_t), \\
    \text{which then gives us } \ & \bm{d} - \bm{a}_t x_t = \frac{1}{\alpha} (\bm{p}_t - \bm{p}_{t+1}).
\end{align*}

Summarizing on both sides and applying the telescoping, we derive \begin{align*}
    \sum_{t=kf}^T (\bm{d} - \bm{a}_t x_t) \leq \frac{1}{\alpha} \sum_{t=kf}^T (\bm{p}_{t} - \bm{p}_{t+1}) = \frac{1}{\alpha} (\bm{p}_{kf} - \bm{p}_{T+1}).
\end{align*}

Then for the positive binding terms, by Lemma \ref{lem:conv-lp} and \eqref{eq:prob-bound}, we have \begin{align} \label{eq:binding}
    \mathbb{E} \Big\lVert \Big(\sum_{t=kf}^T (\bm{d} - \bm{a}_t x_t) \Big)^{B^+} \Big\rVert & \leq \mathbb{E} \Big\lVert \sum_{t=kf}^T (\bm{d} - \bm{a}_t x_t) \Big\rVert \nonumber \\
    & \leq \frac{1}{\alpha} \mathbb{E} \lVert \bm{p}_{kf} - \bm{p}_{T+1} \rVert \nonumber \\
    & \leq \frac{1}{\alpha} \mathbb{E} \Big[\lVert \bm{p}_{kf} - \bm{p}^* \rVert + \lVert \bm{p}_{T+1} - \bm{p}^* \rVert \Big] \nonumber \nonumber \\
    & \leq \frac{C_{lp}}{\sqrt{kf}} \cdot \frac{1}{\alpha} + \frac{C_b}{\sqrt{(T-kf) kf}} \cdot \frac{1}{\alpha \sqrt{\alpha}} + C_b \cdot \frac{1}{\sqrt{\alpha}} + \frac{C_b}{T^2} \cdot \frac{1}{\alpha}. 
\end{align}

Combining the results of \eqref{eq:fo-dual}, Lemma \ref{lem:vio}, and \eqref{eq:binding} together, we obtain the final result as: \begin{align} \label{eq:fo-last-batch}
    \Delta_{T_3} \leq \frac{m(\bar{a}+\bar{d})^2}{\nu \lambda} f \alpha + \Big( \frac{2C_{lp}}{\sqrt{kf}} + \frac{2C_b}{T^2} + \frac{C_{lp} \log(f)}{\nu \lambda kf} \Big) \cdot \frac{1}{\alpha} + \frac{2C_b}{\sqrt{(T-kf) kf}} \cdot \frac{2}{\alpha \sqrt{\alpha}} + \frac{2C_b}{\sqrt{\alpha}}.  
\end{align}

We select the optimal learning rate $\alpha = f^{-2/3}$ to minimize $\Delta_{T_3}$ in \eqref{eq:fo-last-batch}. We prove this result in cases. \begin{enumerate}
    \item If $T - kf \leq f^{1/3}$, the regret must be smaller than the length of the batch, so $\Delta_{T_3} \leq T - kf \leq f^{1/3}$.
    \item If $T - kf > f^{1/3}$, then we use this property to bound the term $\frac{2C_b}{\sqrt{(T-kf) kf}} \cdot \frac{2}{\alpha \sqrt{\alpha}}$. We derive 
    \begin{align} \label{eq:opt-last}
        \Delta_{T_3} & \leq \frac{m(\bar{a}+\bar{d})^2}{\nu \lambda} f \cdot f^{-2/3} + \Big( \frac{2C_{lp}}{\sqrt{kf}} + \frac{2C_b}{T^2} + \frac{C_{lp} \log(f)}{\nu \lambda kf} \Big) \cdot f^{2/3} + \frac{4C_b}{\sqrt{f^{1/3} \cdot kf}} \cdot f + 2C_b \cdot f^{1/3} \nonumber \\
        &= \frac{m(\bar{a}+\bar{d})^2}{\nu \lambda} f^{1/3} + \frac{2C_{lp}}{\sqrt{k}} f^{1/6} + \frac{2C_b}{T} + \frac{C_{lp} \log(f)}{\nu \lambda k} f^{1/3} + \frac{4C_b}{\sqrt{k}} \cdot f^{1/3} + 2C_b \cdot f^{1/3} \nonumber \\
        & \leq \Big(\frac{m(\bar{a}+\bar{d})^2}{\nu \lambda} + \frac{2C_{lp}}{\sqrt{k}} + \frac{C_{lp} \log(f)}{\nu \lambda k} + \frac{4C_b}{\sqrt{k}} + 2C_b \Big) \cdot f^{1/3}. 
    \end{align}
\end{enumerate}
This completes the proof.

\paragraph{Proof of \cref{lem:fo-special}.} According to Lemma \ref{lem:fo-rec}, with $\lVert \bm{p}_0 - \bm{p}^* \rVert \leq f^{-1/3}$ and $\alpha < \frac{1}{\nu \lambda}$, we have \begin{align}\label{eq:spe-cond-rec}
    & \mathbb{E} \big[ \lVert \bm{p}_{f+1} - \bm{p}^* \rVert^2 \mid \bm{p}_{0} \big] \leq \frac{\lVert \bm{p}_0 - \bm{p}^* \rVert^2}{\nu \lambda \alpha f} + \frac{\alpha m(\bar{a} + \bar{d})^2}{\nu \lambda}, \nonumber \\
    & \mathbb{E} \lVert \bm{p}_{f+1} - \bm{p}^* \rVert \leq \frac{1}{f^{1/3} \sqrt{\nu \lambda f}} \cdot \frac{1}{\sqrt{\alpha}} + \frac{m(\bar{a} + \bar{d})}{\sqrt{\nu \lambda}} \cdot \sqrt{\alpha} + \frac{\bar{p}}{T^2}.
\end{align}

Then according to Lemma \ref{lem:vio}, take $C_w = \max\{\frac{1}{\nu \lambda}, \frac{m(\bar{a} + \bar{d})}{\sqrt{\nu \lambda}}, \bar{p}\}$, we derive the constraint violation follows 
\begin{align}\label{eq:spe-vio-dev}
    \mathbb{E} \Big\lVert \Big(\sum_{t=0}^f (\bm{a}_t x_t - \bm{d}) \Big)^+ \Big\rVert & \leq \mathbb{E} \Big\lVert \sum_{t=0}^f (\bm{a}_t x_t - \bm{d}) \Big\rVert \nonumber \\
    & \leq \frac{1}{\alpha} \mathbb{E} \lVert \bm{p}_{f+1} - \bm{p}_{0} \rVert \nonumber \\
    & \leq \frac{1}{\alpha} \mathbb{E} \Big[\lVert \bm{p}_{0} - \bm{p}^* \rVert + \lVert \bm{p}_{f+1} - \bm{p}^* \rVert \Big] \nonumber \\
    & \leq \frac{1}{\alpha} \Big(\frac{1}{f^{1/3}} + C_w \Big[ \frac{1}{f^{1/3} \sqrt{f}} \cdot \frac{1}{\sqrt{\alpha}} + \sqrt{\alpha} + \frac{1}{T^2} \Big] \Big) \nonumber \\
    & \leq \frac{1}{f^{1/3}} \cdot \frac{1}{\alpha} + \frac{C_w}{f^{5/6}} \cdot \frac{1}{\alpha \sqrt{\alpha}} + C_w \cdot \frac{1}{\sqrt{\alpha}} + \frac{C_w}{T^2} \cdot \frac{1}{\alpha}. 
\end{align}

Similar to \eqref{eq:fo-dual} in Theorem \ref{thm:fo-regret-2}, we derive the dual distance as \begin{align} \label{eq:spe-fo-dual}
    \sum_{t=0}^f \mathbb{E} \lVert \bm{p}_t - \bm{p}^* \rVert^2 & \leq \sum_{t=0}^f \Big[ \frac{\lVert \bm{p}_0 - \bm{p}^* \rVert^2}{\nu \lambda \alpha t} + \frac{m(\bar{a}+\bar{d})^2}{\nu \lambda} \alpha \Big] \nonumber \\
    & \leq \frac{\lVert \bm{p}_{0} - \bm{p}^* \rVert^2}{\nu \lambda \alpha} \log(f) + \frac{m(\bar{a}+\bar{d})^2}{\nu \lambda} f \alpha \nonumber \\
    & \leq \frac{\log(f)}{\nu \lambda f^{2/3}} \cdot \frac{1}{\alpha} + \frac{m(\bar{a}+\bar{d})^2}{\nu \lambda} f \alpha. 
\end{align}

Similar to \eqref{eq:binding} in Theorem \ref{thm:fo-regret-2}, we derive the positive projection for binding terms as 
\begin{align} \label{eq:spe-binding}
    \mathbb{E} \Big\lVert \Big(\sum_{t=0}^f (\bm{d} - \bm{a}_t x_t) \Big)^{B^+} \Big\rVert & \leq \mathbb{E} \Big\lVert \sum_{t=0}^f (\bm{d} - \bm{a}_t x_t) \Big\rVert \nonumber \\
    & \leq \frac{1}{\alpha} \mathbb{E} \lVert \bm{p}_{0} - \bm{p}_{f+1} \rVert \nonumber \\
    & \leq \frac{1}{\alpha} \mathbb{E} \Big[\lVert \bm{p}_{0} - \bm{p}^* \rVert + \lVert \bm{p}_{f+1} - \bm{p}^* \rVert \Big] \nonumber \nonumber \\
    & \leq \frac{1}{f^{1/3}} \cdot \frac{1}{\alpha} + \frac{C_w}{f^{1/3} \sqrt{f}} \cdot \frac{1}{\alpha \sqrt{\alpha}} + C_w \cdot \frac{1}{\sqrt{\alpha}} + \frac{C_w}{T^2} \cdot \frac{1}{\alpha}. 
\end{align}

Therefore, combining the results in \eqref{eq:spe-vio-dev}, \eqref{eq:spe-fo-dual}, and \eqref{eq:spe-binding} together, we have \begin{align} \label{eq:spe-fo-first}
    \Delta_{T_1 \text{warm}} & \leq  \frac{\log(f)}{\nu \lambda f^{2/3}} \cdot \frac{1}{\alpha} + \frac{m(\bar{a}+\bar{d})^2}{\nu \lambda} f \alpha \nonumber \\
    & \quad + \frac{2}{f^{1/3}} \cdot \frac{1}{\alpha} + \frac{2C_w}{f^{1/3} \sqrt{f}} \cdot \frac{1}{\alpha \sqrt{\alpha}} + 2C_w \cdot \frac{1}{\sqrt{\alpha}} + \frac{2C_w}{T^2} \cdot \frac{1}{\alpha} \nonumber \\
    & \leq \frac{m(\bar{a}+\bar{d})^2}{\nu \lambda} f \alpha + \Big(\frac{\log(f)}{\nu \lambda f^{2/3}} + \frac{2}{f^{1/3}} + \frac{2C_w}{T^2} \Big) \cdot \frac{1}{\alpha} + \frac{2C_w}{f^{1/3} \sqrt{f}} \cdot \frac{1}{\alpha \sqrt{\alpha}} + 2C_w \cdot \frac{1}{\sqrt{\alpha}}.
\end{align}

Thus, taking the optimal learning rate $\alpha = f^{-2/3}$, we obtain the regret for the first batch with a warm start as: 
\begin{align}\label{eq:warm-fo-1}
    \Delta_{T_1 \text{warm}} & \leq \frac{m(\bar{a}+\bar{d})^2}{\nu \lambda} f^{1/3} + \Big(\frac{\log(f)}{\nu \lambda f^{2/3}} + \frac{2}{f^{1/3}} + \frac{2C_w}{T^2} \Big) \cdot f^{2/3} + \frac{2C_w}{f^{1/3} \sqrt{f}} \cdot f + 2C_w \cdot f^{1/3} \nonumber \\
    & \leq \frac{m(\bar{a}+\bar{d})^2}{\nu \lambda} f^{1/3} + \frac{\log(f)}{\nu \lambda} + 2f^{1/3} + \frac{2C_w}{T} + 2C_w f^{1/6} + 2C_w \cdot f^{1/3} \nonumber \\
    & \leq \Big( \frac{m(\bar{a}+\bar{d})^2}{\nu \lambda} + 4C_w \Big) \cdot f^{1/3} + \log(f).
\end{align}
This completes the proof. 

\section{Main Results}
\label{app:main}

In this section, we demonstrate the proof for all theoretical results that we proposed in the main body of the paper. We instate Assumption \ref{ass:1} and \ref{ass:2} are satisfied.

\subsection{Proof of Theorem \ref{thm:regret}.} 

By Theorem \ref{thm:gap}, we obtain \begin{align*}
    \Delta_T &= \mathbb{E}[r(\bm{x}) + v(\bm{x})] \\
    & \leq \lVert \bm{p}^* \rVert \cdot \mathbb{E} \bigg\lVert \Big(\bm{b} - \sum_{t=1}^T \bm{a}_t x_t \Big)^{B^+} \bigg\rVert + \frac{\mu \bar{a}^2}{2} \sum_{t=1}^T \mathbb{E} \lVert \bm{p}_t - \bm{p}^* \rVert^2 + \mathbb{E} \Big\lVert \Big(\sum_{t=1}^T \bm{a}_t x_t - \bm{b} \Big)^+ \Big\rVert \\
    & \leq \lVert \bm{p}^* \rVert \cdot \mathbb{E} \left[ \lVert (\bm{b} - \bm{Ax})^{B+} \rVert \right] + \mu \bar{a}^2 \sum_{t=1}^T \mathbb{E} \left[ \lVert \bm{p}_t - \bm{p}^* \rVert^2 \right] + \mathbb{E} \left[ \lVert (\bm{Ax} - \bm{b})^+ \rVert \right]. 
\end{align*}
This completes the proof. 

\subsection{Proof of Theorem \ref{thm:spectrum}.}

Combining the results of Theorem \ref{thm:lp-regret}, Theorem \ref{thm:fo-regret-1}, and Theorem \ref{thm:fo-regret-2}, for some constant $C_{reg} > 0$, we obtain: \begin{align}\label{eq:final-reg}
    \Delta_T &= \Delta_{T_1} + \Delta_{T_2} + \Delta_{T_3} \nonumber \\
    & \leq \left(\frac{m(\bar{a} + \bar{d})^2}{2} + \frac{2\bar{r} + m(\bar{a}+\bar{d})^2}{\underline{d}} + m(\bar{a}+\bar{d}) \right) \sqrt{f} \nonumber \\
    & \quad + \Big(C_{lp} + \frac{m C}{\nu^2 \lambda^2} + (\bar{a} + \bar{d} + \delta) \frac{C}{\delta^2} \Big) \log(k) \nonumber \\
    & \quad + \Big(\frac{m(\bar{a}+\bar{d})^2}{\nu \lambda} + \frac{2C_{lp}}{\sqrt{k}} + \frac{C_{lp} \log(f)}{\nu \lambda k} + \frac{4C_b}{\sqrt{k}} + 2C_b \Big) \cdot f^{1/3}. 
\end{align}

Therefore, we achieve the worst-case regret of: \begin{align}\label{eq:final}
    \Delta_T = \mathcal{O} (\log(k) + \sqrt{f} + f^{1/3}).
\end{align}

For special cases, \begin{enumerate}
    \item If we use LP-based method on the first batch, then we will have $\Delta_{T_1} \leq \log(f)$. Total performance follows 
    \begin{align}\label{eq:special-1}
        \Delta_T \leq \log(\max\{f, k\}) + f^{1/3}) \leq \log(\sqrt{T}) + f^{1/3} \in \mathcal{O}(\log(T) + f^{1/3}). 
    \end{align}
    \item If we have a warm start for the first batch, with the initialization $\lVert \bm{p}_0 - \bm{p}^* \rVert \leq f^{-1/3}$, then the first batch regret achieves $\Delta_{T_1} \in \mathcal{O}(f^{1/3} + \log(f))$ by Lemma \ref{lem:fo-special}. Thus, total performance follows 
    \begin{align} \label{eq:special-2}
        \Delta_T \leq \log(\max\{f, k\}) + 2f^{1/3}) \in \mathcal{O}(\log(T) + f^{1/3}). 
    \end{align}
\end{enumerate}
This completes our proof. 

\section{Auxiliary Results} 
\label{app:addexp}

In this section, we provide auxiliary results to support the proof in the previous three sections. These lemmas focus on pure mathematical derivations. 

\subsection{Technical Support for LP-based Analysis}

\begin{lem} \label{aux:dynamic}
    Denote $d_{i, (j+1)f}, d_{i, jf}$ as the average consumption of $i$-th type resource at time $(j+1)f$ and $jf$. There exists a constant $C_{rec}$ depending on $\bar{d}, \bar{a}, m, \nu, \lambda, \mu$, and $C_{lp}$ such that: \begin{align*}
        \mathbb{E} (d_{i, (j+1)f} - d_i)^2 & \leq \mathbb{E} ({d}_{i, jf} - d_i)^2 + \frac{C_{rec}}{(k - j - 1)^2 f} + \frac{4\mu \bar{a}^2 \sqrt{C_{lp}}}{(k - j -1) \sqrt{(j+1)f}} \sqrt{\mathbb{E}[({d}_{i, jf} - d_i)^2]}.
    \end{align*}
\end{lem}

\begin{lem}\label{aux:sum-d}
    With the recursion relation in \eqref{eq:rec}, there exists a constant $C > 0$ depending on $\bar{d}, \bar{a}, m, \nu, \lambda, \mu$, and $C_{lp}$ such that the summation of the total deviation of $\bm{d}_t$ with the original $\bm{d}$ satisfies: \begin{align*}
        \sum_{j=1}^{k} \mathbb{E} \left[ ({d}_{i, jf} - {d}_i)^2 \right] & \leq \frac{C}{f} \log(k).
    \end{align*}
\end{lem}

\paragraph{Proof of \cref{aux:dynamic}.} We analyze each term in \eqref{eq:expand}. The key technique we use is to take conditional expectations and simplify the double summations. 
\begin{enumerate}
    \item[(a)] Term 1.
    \begin{align*}
        & \mathbb{E} \left[\frac{(\sum_{\ell = jf+1}^{(j+1)f} d_{i, jf} - {a}_{i, \ell} I(r_{\ell} > \bm{a}_{\ell}^\top \bm{p}_{jf}^*))^2}{(T - (j+1)f)^2} \right] \\
        = \ & \frac{1}{(T-(j+1)f)^2} \ \mathbb{E} \left[ \sum_{\ell=jf+1}^{(j+1)f} \sum_{s=jf+1}^{(j+1)f} \mathbb{E} \Big[(d_{i, jf} - {a}_{i, \ell} I(r_{\ell} > \bm{a}_{\ell}^\top \bm{p}_{jf}^*)) (d_{i, jf} - {a}_{i, s} I(r_{s} > \bm{a}_{s}^\top \bm{p}_{jf}^*)) \mid \bm{d}_{jf} \Big] \right] \\
        = \ & \frac{1}{(T-(j+1)f)^2} \ \mathbb{E} \left[\sum_{\ell=jf+1}^{(j+1)f} \mathbb{E} \Big[ (d_{i, jf} - {a}_{i, \ell} I(r_{\ell} > \bm{a}_{\ell}^\top \bm{p}_{jf}^*))^2 \mid \bm{d}_{jf} \Big] \right] \\
        & + \frac{1}{(T-(j+1)f)^2} \ \mathbb{E} \left[\sum_{\ell \neq j, \ell=jf+1}^{(j+1)f} \mathbb{E} \Big[ (d_{i, jf} - {a}_{i, \ell} I(r_{\ell} > \bm{a}_{\ell}^\top \bm{p}_{jf}^*)) (d_{i, jf} - {a}_{i, j} I(r_{j} > \bm{a}_{j}^\top \bm{p}_{jf}^*)) \mid \bm{d}_{jf} \Big] \right] \\
        \leq \ & \frac{f(\bar{a} + \bar{d})^2}{(T-(j+1)f)^2} + \frac{\mathbb{E} \left[\sum_{\ell \neq j, \ell=jf+1}^{(j+1)f} \mathbb{E} \Big[ d_{i, jf} - {a}_{i, \ell} I(r_{\ell} > \bm{a}_{\ell}^\top \bm{p}_{jf}^*) | \bm{d}_{jf} \Big] \mathbb{E} \Big[ d_{i, jf} - {a}_{i, j} I(r_{j} > \bm{a}_{j}^\top \bm{p}_{jf}^*) | \bm{d}_{jf} \Big] \right]}{(T-(j+1)f)^2} \\
        = \ & \frac{f(\bar{a} + \bar{d})^2}{(T-(j+1)f)^2} \ \ \ (\text{since } \mathbb{E} \Big[ d_{i, jf} - {a}_{i, \ell} I(r_{\ell} > \bm{a}_{\ell}^\top \bm{p}_{jf}^*) | \bm{d}_{jf} \Big] = 0 \text{ for binding terms}) \\
        \leq \ & \frac{(\bar{a} + \bar{d})^2}{(k - j - 1)^2 f}. \ \ \ (\text{since } \lfloor T \rfloor = k \cdot f)
    \end{align*}
    \item[(b)] Term 2.
    \begin{align*}
        & \mathbb{E} \left[\frac{(\sum_{\ell = jf+1}^{(j+1)f} {a}_{i, \ell} I(r_{\ell} > \bm{a}_{\ell}^\top \bm{p}_{jf}^*) - {a}_{i, \ell} I(r_{\ell} > \bm{a}_{\ell}^\top \bm{p}_{(j+1)f}))^2}{(T - (j+1)f)^2} \right] \\
        = \ & \frac{\mathbb{E} \Bigg[ \sum_{\ell, s =jf+1}^{jk+1)f} \mathbb{E} \Big[({a}_{i, \ell} (I(r_{\ell} > \bm{a}_{\ell}^\top \bm{p}_{jf}^*) - I(r_{\ell} > \bm{a}_{\ell}^\top \bm{p}_{(j+1)f}))) \cdot ({a}_{i, s} (I(r_{s} > \bm{a}_{s}^\top \bm{p}_{jf}^*) - I(r_{s} > \bm{a}_{s}^\top \bm{p}_{(j+1)f}))) \Big| \bm{d}_{jf} \Big] \Bigg]}{(T - (j+1)f)^2}.
    \end{align*}
    When $\ell = s$, \begin{align*}
        \mathbb{E} \Big[({a}_{i, \ell} (I(r_{\ell} > \bm{a}_{\ell}^\top \bm{p}_{jf}^*) - I(r_{\ell} > \bm{a}_{\ell}^\top \bm{p}_{(j+1)f})))^2 \mid \bm{d}_{jf} \Big] \leq \bar{a}^2. 
    \end{align*}
    When $\ell \neq s$, by Assumption \ref{ass:2},  \begin{align*}
        & \mathbb{E} \Big[({a}_{i, \ell} (I(r_{\ell} > \bm{a}_{\ell}^\top \bm{p}_{jf}^*) - I(r_{\ell} > \bm{a}_{\ell}^\top \bm{p}_{(j+1)f}))) ({a}_{i, s} (I(r_{s} > \bm{a}_{s}^\top \bm{p}_{jf}^*) - I(r_{s} > \bm{a}_{s}^\top \bm{p}_{(j+1)f}))) \mid \bm{d}_{jf} \Big] \\
        = \ & \mathbb{E} \Bigg[ \mathbb{E} \Big[({a}_{i, \ell} (I(r_{\ell} > \bm{a}_{\ell}^\top \bm{p}_{jf}^*) - I(r_{\ell} > \bm{a}_{\ell}^\top \bm{p}_{(j+1)f}))) ({a}_{i, s} (I(r_{s} > \bm{a}_{s}^\top \bm{p}_{jf}^*) - I(r_{s} > \bm{a}_{s}^\top \bm{p}_{(j+1)f}))) \mid \bm{a}_i, \bm{a}_s \Big] \mid \bm{d}_{jf} \Bigg] \\
        = \ & \mathbb{E} \Bigg[ {a}_{i, \ell} \mathbb{E} \Big[ I(r_{\ell} > \bm{a}_{\ell}^\top \bm{p}_{jf}^*) - I(r_{\ell} > \bm{a}_{\ell}^\top \bm{p}_{(j+1)f}) \mid \bm{a}_{\ell} \Big] \cdot {a}_{i, s} \mathbb{E} \Big[ I(r_{s} > \bm{a}_{s}^\top \bm{p}_{jf}^*) - I(r_{s} > \bm{a}_{s}^\top \bm{p}_{(j+1)f}) \mid \bm{a}_s \Big] \mid \bm{d}_{jf} \Bigg] \\
        = \ & \mathbb{E} \Bigg[ {a}_{i, \ell} (P(r_{\ell} > \bm{a}_{\ell}^\top \bm{p}_{jf}^* \mid \bm{a}_{\ell}) - P(r_{\ell} > \bm{a}_{\ell}^\top \bm{p}_{(j+1)f} \mid \bm{a}_{\ell})) \cdot {a}_{i, s} (P(r_{s} > \bm{a}_{s}^\top \bm{p}_{jf}^* \mid \bm{a}_s) - P(r_{s} > \bm{a}_{s}^\top \bm{p}_{(j+1)f} \mid \bm{a}_s)) \mid \bm{d}_{jf} \Bigg] \\
        \leq \ & \mathbb{E} \Bigg[ \mu {a}_{i, \ell} \bm{a}_{\ell}^\top (\bm{p}_{(j+1)f} - \bm{p}_{jf}^*) \cdot \mu {a}_{i, s} \bm{a}_s^\top (\bm{p}_{(j+1)f} - \bm{p}_{jf}^*) \mid \bm{d}_{jf} \Bigg] \ \ \text{ (using Assumption here) } \\
        \leq \ & \mu^2 \bar{a}^4 \ \mathbb{E} \big[(\bm{p}_{(j+1)f} - \bm{p}_{jf}^*)^2 \mid \bm{d}_{jf} \big]. 
    \end{align*}
    Combining these two cases together and by the convergence of LP-based method, we obtain the bound for Term 2 as \begin{align*}
        & \mathbb{E} \left[\frac{(\sum_{\ell = jf+1}^{(j+1)f} {a}_{i, \ell} I(r_{\ell} > \bm{a}_{\ell}^\top \bm{p}_{jf}^*) - {a}_{i, \ell} I(r_{\ell} > \bm{a}_{\ell}^\top \bm{p}_{(j+1)f}))^2}{(T - (j+1)f)^2} \right] \\
        \leq \ & \frac{1}{(T-(j+1)f)^2} \ \mathbb{E} \left[\sum_{\ell = jf+1}^{(j+1)f} \bar{a}^2 + \sum_{\ell \neq s, \ell = jf+1}^{(j+1)f} \mu^2 \bar{a}^4 \ \mathbb{E} \big[(\bm{p}_{(j+1)f} - \bm{p}_{jf}^*)^2 \big| \bm{d}_{jf} \big] \Big| \bm{d}_{jf} \right] \\
        \leq \ & \frac{1}{(T-(j+1)f)^2} \ \Big( f\bar{a}^2 + f^2 \mu^2 \bar{a}^4 \mathbb{E}[(\bm{p}_{(j+1)f} - \bm{p}_{jf}^*)^2 \big| \bm{d}_{jf}] \Big) \\
        \leq \ & \frac{1}{(T-(j+1)f)^2} \ \Big( f\bar{a}^2 + f^2 \mu^2 \bar{a}^4 \frac{C_{lp}}{jf} \Big) \ \ \text{ (by Lemma \ref{lem:conv-lp})} \\
        \leq \ & \frac{\bar{a}^2 + \frac{1}{j} \mu^2 \bar{a}^4 C_{lp}}{(k - j - 1)^2 f}.
    \end{align*}
    \item[(c)] Term 3. 
    \begin{align*}
        & 2\mathbb{E} \left[({d}_{i, jf} - d_i) \left(\frac{\sum_{\ell = jf+1}^{(j+1)f} {d}_{i, jf} - {a}_{i, \ell} I(r_{\ell} > \bm{a}_{\ell}^\top \bm{p}_{jf}^*)}{T - (j+1)f} \right) \right] \\
        = \ & \frac{2}{T-(j+1)f} \ \mathbb{E} \left[ \sum_{\ell = jf+1}^{(j+1)f} \mathbb{E} \Big[{d}_{i, jf} - {a}_{i, \ell} I(r_{\ell} > \bm{a}_{\ell}^\top \bm{p}_{jf}^*) \mid \bm{d}_{jf} \Big] \cdot ({d}_{i, jf} - d_i) \right] \\
        = \ & 0. \ \ \ \text{ (by the definition of binding terms)}
    \end{align*}
    \item[(d)] Term 4. 
    \begin{align*}
        & 2\mathbb{E} \left[({d}_{i, jf} - d_i) \left(\frac{\sum_{\ell = jf+1}^{(j+1)f} {a}_{i, \ell} I(r_{\ell} > \bm{a}_{\ell}^\top \bm{p}_{jf}^*) - {a}_{i, \ell} I(r_{\ell} > \bm{a}_{\ell}^\top \bm{p}_{(j+1)f})}{T - (j+1)f} \right) \right] \\
        = \ & \frac{2}{T-(j+1)f} \ \mathbb{E} \Bigg[\sum_{\ell = jf+1}^{(j+1)f} ({d}_{i, jf} - {d}_i) {a}_{i, \ell} \ \mathbb{E} \Big[I(r_{\ell} > \bm{a}_{\ell}^\top \bm{p}_{jf}^*) - I(r_{\ell} > \bm{a}_{\ell}^\top \bm{p}_{(j+1)f}) \mid \bm{a}_{\ell} \Big] \Bigg] \\
        \leq \ & \frac{2}{T-(j+1)f} \sum_{\ell = jf+1}^{(j+1)f} \mathbb{E} \Bigg[ ({d}_{i, jf} - {d}_i) {a}_{i, \ell} \ \Big[P(r_{\ell} > \bm{a}_{\ell}^\top \bm{p}_{jf}^* \mid \bm{a}_{\ell}) - P(r_{\ell} > \bm{a}_{\ell}^\top \bm{p}_{(j+1)f} \mid \bm{a}_{\ell}) \Big] \Bigg] \\
        \leq \ & \frac{2}{T-(j+1)f} \sum_{\ell = jf+1}^{(j+1)f} \mathbb{E} \Bigg[ ({d}_{i, jf} - {d}_i) \mu {a}_{i, \ell} \bm{a}_{\ell}^\top (\bm{p}_{(j+1)f} - \bm{p}_{jf}^*) \Bigg] \ \ \text{ (by Assumption \ref{ass:2})} \\
        \leq \ & \frac{2\mu \bar{a}^2}{T-(j+1)f} \sum_{\ell = jf+1}^{(j+1)f} \sqrt{\mathbb{E}[({d}_{i, jf} - {d}_i)^2]} \cdot  \sqrt{\mathbb{E} [(\bm{p}_{(j+1)f} - \bm{p}_{jf}^*)^2]} \ \ \ \text{(by Cauchy's inequality)} \\
        \leq \ & \frac{2\mu \bar{a}^2}{T-(j+1)f} \cdot  \frac{\sqrt{C_{lp}} f}{\sqrt{jf}} \sqrt{\mathbb{E}[({d}_{i, jf} - {d}_i)^2]} \\
        = \ & \frac{4\mu \bar{a}^2 \sqrt{C_{lp}}}{(k - j -1) \sqrt{(j+1)f}} \sqrt{\mathbb{E}[({d}_{i, jf} - {d}_i)^2]}.
    \end{align*}
    \item[(e)] Term 5. 
    \begin{align*}
        & 2\mathbb{E} \left(\frac{\sum_{\ell = jf+1}^{(j+1)f} {d}_{i, jf} - {a}_{i, \ell} I(r_{\ell} > \bm{a}_{\ell}^\top \bm{p}_{jf}^*)}{T - (j+1)f} \cdot \frac{\sum_{\ell = jf+1}^{(j+1)f} {a}_{i, \ell} I(r_{\ell} > \bm{a}_{\ell}^\top \bm{p}_{jf}^*) - {a}_{i, \ell} I(r_{\ell} > \bm{a}_{\ell}^\top \bm{p}_{(j+1)f})}{T - (j+1)f} I(jf < \tau) \right) \\
        \leq \ & 2 \sqrt{\mathbb{E} \left(\frac{\sum_{\ell = jf+1}^{(j+1)f} {d}_{i, jf} - {a}_{\ell} I(r_{\ell} > \bm{a}_{\ell}^\top \bm{p}_{jf}^*)}{T - (j+1)f} \right)^2} \sqrt{\mathbb{E} \left(\frac{\sum_{\ell = jf+1}^{(j+1)f} {a}_{i, \ell} I(r_{\ell} > \bm{a}_{\ell}^\top \bm{p}_{jf}^*) - {a}_{i, \ell} I(r_{\ell} > \bm{a}_{\ell}^\top \bm{p}_{(j+1)f})}{T - (j+1)f} \right)^2} \\
        \leq \ & 2 \sqrt{\frac{(\bar{a}+\bar{d})^2}{(k - j -1)^2 f}} \cdot \sqrt{\frac{\bar{a}^2 + \frac{1}{j} \mu^2 \bar{a}^4 C_{lp}}{(k - j - 1)^2 f}} \ \ \ \text{ (by results of Term 1 and Term 2)} \\
        = \ & \frac{2\bar{a}(\bar{a}+\bar{d}) \sqrt{1 + \frac{1}{j} \mu^2 \bar{a}^2 C_{lp}}}{(k-j-1)^2 f}.
    \end{align*}
\end{enumerate} 

Combining all the terms, we obtain the upper bound as:
\begin{align*}
    \mathbb{E} \left[ ({d}_{i, (j+1)f} - {d}_i)^2 \right] & \leq \mathbb{E} ({d}_{i, jf} - {d}_i)^2 + \frac{(\bar{a} + \bar{d})^2}{(k - j - 1)^2 f} + \frac{\bar{a}^2 + \frac{1}{j} \mu^2 \bar{a}^4 C_{lp}}{(k - j - 1)^2 f} \\
    & \ \ + \frac{2\bar{a}(\bar{a}+\bar{d}) \sqrt{1 + \frac{1}{j} \mu^2 \bar{a}^2 C_{lp}}}{(k - j - 1)^2 f} + \frac{4\mu \bar{a}^2 \sqrt{C_{lp}}}{(k - j -1) \sqrt{(j+1)f}} \sqrt{\mathbb{E}[({d}_{i, jf} - {d}_i)^2]}.
\end{align*}

Taking $C_{rec} = (\bar{a} + \bar{d})^2 + \bar{a}^2 + \mu^2 \bar{a}^4 C_{lp} + 2\bar{a}(\bar{a}+\bar{d}) \sqrt{1 + \mu^2 \bar{a}^2 C_{lp}}$ completes the proof. 

\paragraph{Proof of \cref{aux:sum-d}.} We consider a general sequence $\{z_j\}_{j=1}^k$ with \begin{align*}
    z_{j+1} \leq z_j + \frac{R}{(k-j-1)^2 f} + \frac{\sqrt{R} \sqrt{z_j}}{(k-j-1) \sqrt{(j+1)f}}
\end{align*}
where $R > 0$ is a constant. 

Taking sum on both sides of the inequality and re-arranging, we have
\begin{align*}
    \sum_{j=1}^{k} (k - j + 1) (z_{j+1} - z_j) & \leq \sum_{j=1}^{k} \frac{16R}{(k-j+1)f} + \sqrt{\frac{16R}{f}} \sum_{j=1}^{k} \frac{\sqrt{z_j}}{\sqrt{j+1}} \\
    & \leq \frac{16R}{f} \log(k) + \sqrt{\frac{16R}{f}} \sum_{j=1}^{k} \frac{\sqrt{z_j}}{\sqrt{j+1}}. 
\end{align*}

Noticing that $\sum_{j=1}^{k} (k - j + 1) (z_{j+1} - z_j) = \sum_{j=1}^{k} z_j$, we have
\begin{align*}
    \frac{16R}{f} \Big(\sum_{j=1}^{k} \frac{1}{j+1} \Big) \cdot \Big(\sum_{j=1}^{k} z_j \Big) & \geq \left(\sqrt{\frac{16R}{f}} \sum_{j=1}^{k} \frac{\sqrt{z_j}}{\sqrt{j+1}} \right)^2 \\
    & \geq \bigg(\sum_{j=1}^{k} z_j - \frac{4R}{f} \log(k)\bigg)^2.
\end{align*}

We treat $\sum_{j=1}^k z_j$ as the variable, then solve and get \begin{align*}
    \sum_{j=1}^{k} z_j \leq \frac{12R}{f} \log(k).
\end{align*}

With this result, consider our recursion in \eqref{eq:rec}, take the constant $C = 12 \max\{ C_{rec}, 16\mu^2 \bar{a}^4 C_{lp}\} > 0$, we obtain: \begin{align*}
    \sum_{j=1}^{k} \mathbb{E} \left[ ({d}_{i, jf} - {d}_i)^2 \right] & \leq \frac{C}{f} \log(k).
\end{align*}

\subsection{Technical Support for First-order Analysis}

\begin{lem}\label{lem:fo-rec}
    Following the updated rule of first-order method, we derive the last dual price satisfies: \begin{align*}
        \mathbb{E} \lVert \bm{p}_{T+1} - \bm{p}^* \rVert \leq \frac{C_{lp}}{\sqrt{\nu \lambda (T-kf) kf}} \cdot \frac{1}{\sqrt{\alpha}} + \frac{m(\bar{a} + \bar{d})}{\sqrt{\nu \lambda}} \cdot \sqrt{\alpha} + \frac{\bar{p}}{T^2}. 
    \end{align*}
\end{lem}

\paragraph{Proof of \cref{lem:fo-rec}.} According to Lemma \ref{lem:conv-fo}, take $\alpha_t = \alpha < \frac{1}{\nu \lambda}$, we derive the conditional expectation as \begin{align} \label{eq:cond-rec}
    \mathbb{E} \big[ \lVert \bm{p}_{T+1} - \bm{p}^* \rVert^2 \mid \bm{p}_{kf} \big] & \leq (1 - \nu \lambda \alpha) \mathbb{E} \big[ \lVert \bm{p}_{T} - \bm{p}^* \rVert^2 \mid \bm{p}_{kf} \big] + \alpha^2 m(\bar{a} + \bar{d})^2 \nonumber \\
    & \leq (1 - \nu \lambda \alpha)^{T-kf} \lVert \bm{p}_{kf} - \bm{p}^* \rVert^2 + \sum_{j=0}^{T-kf-1} \alpha^2 m(\bar{a} + \bar{d})^2 (1 - \nu \lambda \alpha)^j \nonumber \\
    & \leq (1 - \nu \lambda \alpha)^{T-kf} \lVert \bm{p}_{kf} - \bm{p}^* \rVert^2 + \frac{\alpha^2 m(\bar{a} + \bar{d})^2}{\nu \lambda \alpha} \nonumber \\
    & \leq \frac{1}{\nu \lambda \alpha (T-kf)} \lVert \bm{p}_{kf} - \bm{p}^* \rVert^2 + \frac{\alpha m(\bar{a} + \bar{d})^2}{\nu \lambda}
\end{align}
where we use the technique of $\sum_{j=0}^{T-kf-1} \alpha^2 m(\bar{a} + \bar{d})^2 (1 - \nu \lambda \alpha)^j \leq \frac{1 - (1 - \nu \lambda \alpha)^{T-kf}}{\nu \lambda \alpha} \leq \frac{1}{\nu \lambda \alpha}$ and $(1 - \nu \lambda \alpha)^{T-kf} \leq \frac{1}{1 + \nu \lambda \alpha (T-kf)} \leq \frac{1}{\nu \lambda \alpha (T-kf)}$. \\

By LP-convergence result in Lemma \ref{lem:conv-lp}, we know $\mathbb{E} \lVert \bm{p}_{kf} - \bm{p}^* \rVert \leq \frac{C_{lp}}{\sqrt{kf}}$. By \textit{Proposition 3.3} in \cite{gao2024decoupling}, we know the event $E \assign \lVert \bm{p}_{kf} - \bm{p}^* \rVert \leq \frac{C_{lp}}{\sqrt{kf}}$ with probability $\mathbb{P} \geq 1 - \frac{1}{T^4}$. By Lemma \ref{lem:dual-bound}, we know $\lVert \bm{p}_t \rVert \leq \bar{p}$. \\

Thus, we have \begin{align}\label{eq:true-rec}
    \mathbb{E} \lVert \bm{p}_{T+1} - \bm{p}^* \rVert^2 & \leq \mathbb{E} \big[ \lVert \bm{p}_{T+1} - \bm{p}^* \rVert^2 | E \big] \cdot \mathbb{P}(E) + \mathbb{E} \big[ \lVert \bm{p}_{T+1} - \bm{p}^* \rVert^2 | {\bar{E}} \big] \cdot \mathbb{P}(\bar{E}) \nonumber \\
    & \leq \frac{C_{lp}}{\nu \lambda (T-kf) kf} \cdot \frac{1}{\alpha} + \frac{m(\bar{a} + \bar{d})^2}{\nu \lambda} \cdot \alpha + \frac{\bar{p}}{T^4}.
\end{align}

Thus, as $\sqrt{a + b} \leq \sqrt{a} + \sqrt{b}$ for any $a, b > 0$, by \eqref{eq:true-rec}, we have \begin{align*}
    \mathbb{E} \lVert \bm{p}_{T+1} - \bm{p}^* \rVert & \leq \frac{C_{lp}}{\sqrt{\nu \lambda (T-kf) kf}} \cdot \frac{1}{\sqrt{\alpha}} + \frac{m(\bar{a} + \bar{d})}{\sqrt{\nu \lambda}} \cdot \sqrt{\alpha} + \frac{\bar{p}}{T^2}. 
\end{align*} 
This completes the proof. 

\section{Supplementary Experiments}
\label{app: supp_exp}

In this section, we provide more experiments to further evaluate the performance of our algorithms. We consider a more general and complex input distribution and include additional comparisons with recent methods. 

\subsection{New Distribution} \label{supp_exp:new}

As an extension of Section \ref{exp:1}, our goal is to evaluate the main algorithms with different choices of re-solving frequency $f \in \{T^{1/3}, T^{1/2}, T^{2/3}\}$. 
We consider a more complex distribution for reward and resource consumption requests and guarantee that Assumptions \ref{ass:1} and \ref{ass:2} are still satisfied. 

Consider 
\begin{align*}
    \text{Input III: } & \ a_{it} \sim \min(1, \max(0, 1 + z)), \ r_t \sim \text{Unif}[0, 1] \\
    & \text{ where } z \sim t(1): \text{Student's t-distribution with 1 degree of freedom}.
\end{align*}
We generate $\{r_t, \bm{a}_t\}_{t=1}^T$ from Input III and keep other parameters $T \in [10^2, 10^6]$ and $d_i \sim \text{Uniform}[1/3, 2/3]$ the same in Section \ref{exp:1}. We report the average result over 100 trials for each experiment and use the classic first-order method with $\mathcal{O}(T^{1/2})$ regret (\cref{alg:first-order}) as a baseline. 

\begin{figure*}[h]
    \vskip -0.1in
    \centering
    \begin{minipage}{0.45\textwidth}
        \centering
        \addtocounter{table}{1}
        \captionsetup{type=table}
        \caption{Algorithms under New Distribution.
        \label{supp_table:1}}
        \vskip 0.06in
        \resizebox{\textwidth}{!}{
        \begin{tabular}{c | c c c c c}
            \toprule
            & $T$ & First-Order & Low freq & Mid freq & High freq \\
            \midrule
            \multirow{5}{*}{Algorithm \textcolor{red}{1}} 
            & $10^2$ & 3.78 & 1.68 & 1.26 & \textbf{1.20} \\
            & $10^3$ & 10.69 & 2.60 & 2.10 & \textbf{1.54} \\
            & $10^4$ & 36.79 & 4.79 & 3.67 & \textbf{2.54} \\
            & $10^5$ & 117.13 & 10.45 & 6.17 & \textbf{3.56} \\
            & $10^6$ & 377.84 & 34.54 & 14.30 & \textbf{5.48} \\
            \midrule
            \multirow{5}{*}{Algorithm \textcolor{red}{2}} 
            & $10^2$ & 3.78 & 1.24 & 1.17 & \textbf{1.13} \\
            & $10^3$ & 10.69 & 2.30 & 1.96 & \textbf{1.52} \\
            & $10^4$ & 36.79 & 4.12 & 2.99 & \textbf{2.15} \\
            & $10^5$ & 117.13 & 9.78 & 4.88 & \textbf{3.05} \\
            & $10^6$ & 377.84 & 25.96 & 8.90 & \textbf{4.86} \\
            \bottomrule
        \end{tabular}
        }
    \end{minipage}
    \quad \quad
    \begin{minipage}{0.5\textwidth}
        \vskip 0.25in
        \centering
        \addtocounter{figure}{1}
        \caption{Regret for various re-solving frequencies.
        \label{supp_fig:1}}
        \resizebox{\textwidth}{!}{
        \includegraphics{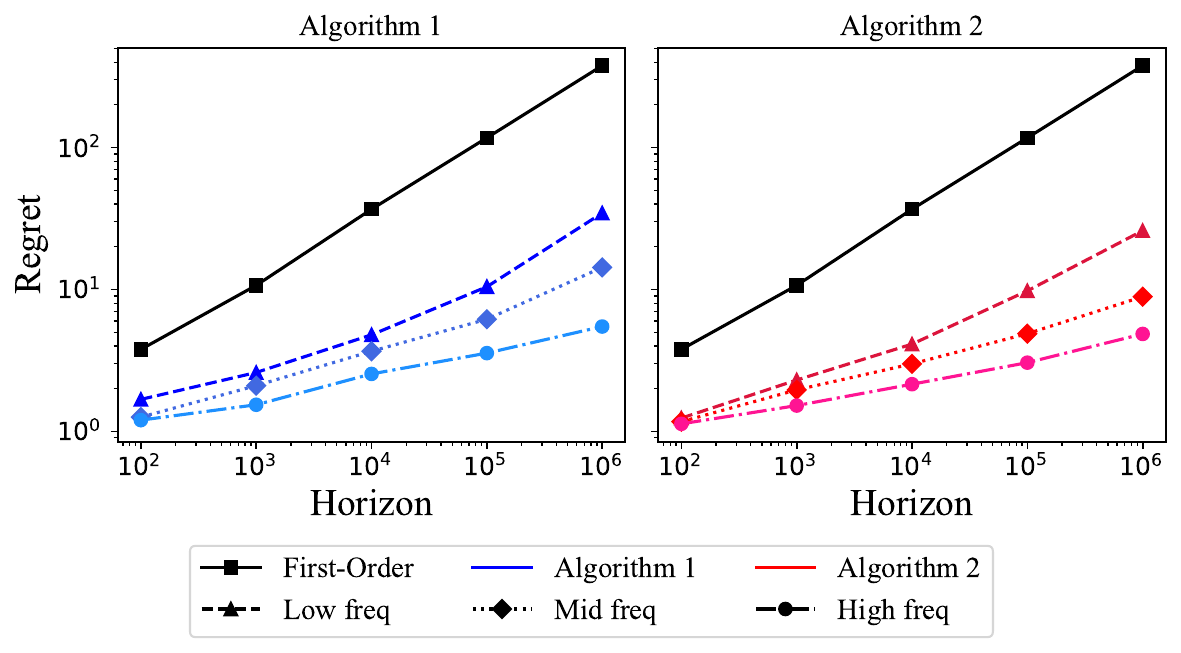}
        }
    \end{minipage}
\end{figure*}

Under the new distribution, our algorithms still exhibit a strong regret performance. As shown in Table \ref{supp_table:1} and Figure \ref{supp_fig:1}, we observe that regret decreases as the re-solving frequency increases. This trend holds consistently across both algorithms and is consistent with the guarantees of Theorem \ref{thm:spectrum}. 

\subsection{More Comparison} \label{supp_exp:com}

Building on the comparison of baseline methods in Section \ref{exp:2}, we evaluate our algorithms against recent works. As a reminder, our algorithms employ frequent LP-solving to learn online dual price under continuous support. \citet{infrequent2024} considers a similar problem using infrequent LP-solving to update the dual variable but under finite support. 

\begin{figure*}[h]
    \centering
    \begin{minipage}{0.38\textwidth}
        \centering
        \captionsetup{type=table}
        \caption{Algorithms comparison.
        \label{supp_table:2}}
        \resizebox{\textwidth}{!}{
        \begin{tabular}{crrr}
            \toprule
            $T$ & Regret & Algorithm & Compute Time (s) \\
            \midrule
            \multirow{4}{*}{$10^3$} & 11.92 & First-Order & 0.002 \\
            & 11.12 & Infrequent LP-based & 0.724 \\
            & \textbf{3.77} & Algorithm \textcolor{red}{1} & \textbf{0.790} \\
            & \textbf{3.69} & Algorithm \textcolor{red}{2} & \textbf{0.782} \\
            \midrule
            \multirow{4}{*}{$10^4$} & 36.37 & First-Order & 0.01 \\
            & 14.20 & Infrequent LP-based & 0.827 \\
            & \textbf{6.69} & Algorithm \textcolor{red}{1} & \textbf{3.015} \\
            & \textbf{6.35} & Algorithm \textcolor{red}{2} & \textbf{3.112} \\
            \midrule
            \multirow{4}{*}{$10^5$} & 110.22 & First-Order & 0.109 \\
            & 20.90 & Infrequent LP-based & 1.028 \\
            & \textbf{16.41} & Algorithm \textcolor{red}{1} & \textbf{52.95} \\
            & \textbf{10.84} & Algorithm \textcolor{red}{2} & \textbf{52.39} \\
            \midrule
            \multirow{4}{*}{$10^6$} & 312.26 & First-Order & 1.106 \\
            & 28.80 & Infrequent LP-based & 4.929 \\
            & \textbf{20.58} & Algorithm \textcolor{red}{1} & \textbf{1261.0} \\
            & \textbf{14.30} & Algorithm \textcolor{red}{2} & \textbf{1305.4} \\
            \bottomrule
        \end{tabular}
        }
    \end{minipage}
    \quad \quad
    \begin{minipage}{0.42\textwidth}
        \vskip 0.1in
        \centering
        \caption{Regret for various algorithms.
        \label{supp_fig:2}}
        \vskip 0.09in
        \resizebox{\textwidth}{!}{
            \includegraphics{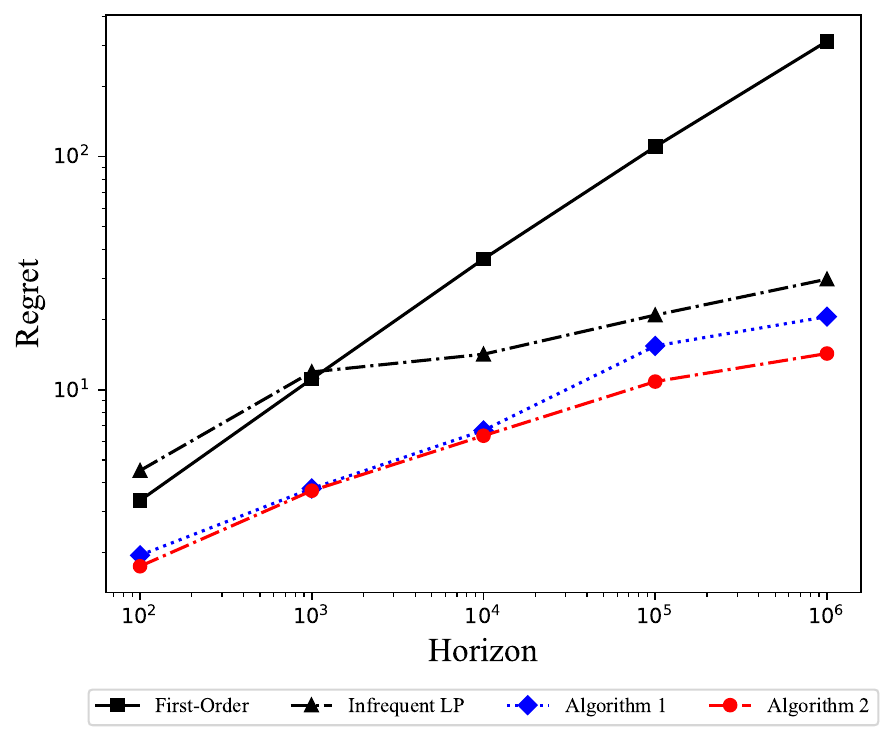}
        }
    \end{minipage}
    \vskip 0.15in
\end{figure*}

To compare them, we adapt our algorithms to finite support and take the re-solving frequency $f = T^{1/3}$ from Section \ref{exp:1}; We take the best-performed parameters for the infrequent method where the number of customer types $n = 50$ and $\alpha = \beta = 0.95$ which control the solving times near start and end. We take the horizon over $T \in [10^2, 10^6]$ and report the average result over 100 trials for each experiment. We still use the first-order method in \cref{alg:first-order} as a baseline. 

Table \ref{supp_table:2} and Figure \ref{supp_fig:2} demonstrate the algorithm regret and computation time across different horizons. While the infrequent LP-based method has faster computation, our algorithms show a competitive performance in decision optimality and achieve lower regret under finite support.


\end{document}